\documentclass[conference]{IEEEtran}
\IEEEoverridecommandlockouts
\usepackage{cite}
\usepackage{amsmath,amssymb,amsfonts}
\usepackage{algorithmic}
\usepackage{graphicx}
\usepackage{textcomp}
\usepackage{xcolor}
\usepackage{hyperref}
\usepackage{orcidlink}
\def\BibTeX{{\rm B\kern-.05em{\sc i\kern-.025em b}\kern-.08em
    T\kern-.1667em\lower.7ex\hbox{E}\kern-.125emX}}
\begin{document}

\title{Online parameter estimation
	for the Crazyflie quadcopter through an EM algorithm
}

\author{\IEEEauthorblockN{Yanhua Zhao \orcidlink{0000-0002-1848-6550}}
}

\maketitle

\begin{abstract}
Drones are becoming more and more popular nowadays. They are small in size, low in cost, and reliable in operation. They contain a variety of sensors and can perform a variety of flight tasks, reaching places that are difficult or inaccessible for humans. Earthquakes damage a lot of infrastructure, making it impossible for rescuers to reach some areas. But drones can help. Many amateur and professional photographers like to use drones for aerial photography. Drones play a non-negligible role in agriculture and transportation too. Drones can be used to spray pesticides, and they can also transport supplies. A quadcopter is a four-rotor drone and has been studied in this paper.\\

In this paper, random noise is added to the quadcopter system  and its effects on the drone system are studied. An extended Kalman filter has been used to estimate the state based on noisy observations from the sensor. Based on a SDE system, a linear quadratic Gaussian controller has been implemented. The expectation maximization algorithm has been applied for parameter estimation of the quadcopter.\\
The results of offline parameter estimation and online parameter estimation are presented. The results show that the online parameter estimation has a slightly larger range of convergence values than the offline parameter estimation.\\
\end{abstract}

\begin{IEEEkeywords}
Quadcopter, Bayesian, EKF, LQG, RST, EM.
\end{IEEEkeywords}

\section{Introduction}
\subsection{Applications of drones}
Many people are already familiar with drones. Drones have gradually integrated into every aspect of our daily life.\\
In the earliest days, drones were used in military reconnaissance. However, with the development of technology, various chips and electronic accessories can be obtained at extremely low prices, so the demand for civilian drones has greatly increased. Now drones have been applied in many fields such as express transportation, aerial photography, agriculture, disaster rescue, mapping and so on. Companies such as Amazon have experimented with practical drone delivery. If the drone delivery technology is mature, this will greatly reduce labor costs. In the face of natural disasters, drones can be used to conduct disaster surveys and transmit high-definition images in real time so that rescuers and  experts can assess risks and develop rescue strategies. In addition, drones can be loaded with a communication base station module and be transformed into a mobile communication base station to provide timely communication services for areas where the communication base station is damaged. Aerial photography enthusiasts can also mount cameras on the drone to take some wonderful videos and pictures. Drones can be used in agriculture, such as crop monitoring, planting and crop spraying. \\
In short, drones are a promising tool and research project, and the economic benefits they bring can not be underestimated.
\subsection{Advantanges and challenges}
There are many types of drones, including four-rotor drones which are also called quadcopters, six-rotor drones, eight-rotor drones, etc. The quadcopter is studied in this paper. The quadopter has a simple mechanical design, the component cost is low, and it is relatively easy to control. The four rotors consist of four electric motors. Adjacent propellers rotate in opposite directions. However, its disadvantage is that the mission load is very small. If you want to increase the load, the size of the aircraft will increase, and the weight of the motor, battery, etc. will also increase. This is a difficult aspect to balance. The wind resistance of quadcopters also needs to be increased.
\subsection{Outline}

In practice, the force acting on a quadcopter is uncertain. All uncertain factors have a non negligible impact on quadcopter systems. Thus, random noise is taken into account and a dynamic model based on the SDE model is proposed. Paper also provides other background informations which include hidden Markov models, differential flatness, different types of sensors, state space models and an expectation maximization algorithm.\\
Due to the presence of random noise, an extended Kalman filter is designed to estimate the state of the quadcopter. The estimator filters out noise and recovers the state based on observations from the sensor.\\ 
The LQG controller is applied to the quadcopter to calculate the trust and torque based on reference trajectory. \\ 
The main goal of the paper is to do parameter estimation for the quadcopter. Because model based control is efficient but requires knowledge of system parameters. Experimental measurement of parameters requires specific gadgets and is laborious and may be inaccurate. If the quadcopter carries some items, its mass and inertia matrix will change accordingly. Therefore, the goal of this project is to learn parameters directly from flight data. In order to be able to achieve stable trajectories with unknown trajectories, an online em algorithm is proposed.\\
At the end, conclusions and future works are presented.

\section{Background}
\subsection{Hidden Markov model}
\begin{figure}[h]
	\centering
	\includegraphics[width=8cm,height=3.8cm]{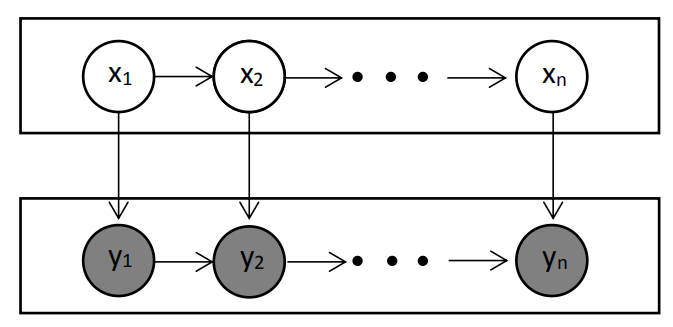}
	\caption{Hidden Markov model}
\end{figure}
Hidden Markov models are used to describe Markov processes with hidden unknown parameters. The real state is invisible to the observer. The observer can only observe observations which contain noise. Therefore, it is assumed that $X$ is the set of all hidden states and $Y$ is the set of all observation states.\\
\begin{equation}
X = [X_{1}, X_{2},..., X_{n}]
\end{equation}
\begin{equation}
Y = [Y_{1}, Y_{2},..., Y_{n}]
\end{equation}
According to the Markov chain rule, it is assumed that the current hidden state depends only on the previous hidden state. This assumption makes the model simpler. The state transition probability of the HMM can be expressed as $p(x_{k}| x_{k-1})$.\\
For the observations, it is assumed that the current observation state depends only on the current hidden state. So the probability for $y_{k}$ given $x_{k}$ is expressed as $p(y_{k}|x_{k})$. \\
In addition, joint probability is defined based on specific models. The joint probability can be expressed as $p(x_{1},...,x_{n}, y_{1},...,y_{n})$.
\subsection{State space models} \label{ssm}
Now a state space model will be presented. There are random variables $X_{1}, X_{2},...,X_{n}$. The current state only depends on the previous state and random noise. The general form is:\\
\begin{equation}
\begin{split}
X_{k} = f(X_{k-1}) + \epsilon_{k}, \ \epsilon_{k} \sim N(0,Q) 
\end{split}
\end{equation}
$\epsilon_{k}$ is Gaussian white noise, also called process noise. $Q$ is the covariance of the process noise. Because of the random noise, the precise value of the state can not be obtained. The noisy observations $Y_{1}, Y_{2}, ... , Y_{k}$ can be obtained. The measurement model connects observation and state. The general form is:\\
\begin{equation}
\begin{aligned}
Y_{k} = h(X_{k}) + \delta_{k}, \  \delta_{k} \sim N(0,R) \label{state2}
\end{aligned}
\end{equation}
$\delta_{k}$ is observation noise and obeys Gaussian distribution. $R$ is the covariance of the observation noise. \\
The conditional probability of $x_{k}$ based on $x_{k-1}$ is:\\ 
\begin{equation} 
\begin{aligned}
p(x_{k}|x_{k-1}) &= (2\pi |Q|)^{-\frac{1}{2}} \\
& \exp\left\{-\frac{1}{2}[x_{k} - f(x_{k-1})]^{T} Q^{-1} [x_{k} - f(x_{k-1})]\right\} 
\end{aligned}
\end{equation}

The conditional probability of $y_{k}$ based on $x_{k}$ is as follows:\\
\begin{equation}
\begin{aligned}
p(y_{k}|x_{k}) &=(2\pi |R|)^{-\frac{1}{2}} \\
&\exp\left\{-\frac{1}{2}[y_{k} - h(x_{k-1})]^{T} R^{-1} [y_{k} - h(x_{k-1})]\right\}
\end{aligned}
\end{equation}
Based on the above conditional probability, the joint probability of all variables can be constructed as following:\\
\begin{equation}
p(x_{1},...,x_{n}, y_{1},...,y_{n}) = p(x_{1})\prod_{i=2}^{N} p(x_{i}|x_{i-1}) \prod_{i=1}^{N} p(y_{i}|x_{i}))\label{HHM}
\end{equation}
The above state space model is a typical hidden Markov model.

\subsection{Differential flatness}
In this section, differential flatness is introduced. Let $\ddot{x}$ be the acceleration of the center of mass and $\ddot{x} = [ acc_{x}, acc_{y}, acc_{z}]$. According to the paper by Ferrin \cite{term1} , let $ t = [acc_{x} , acc_{y}, acc_{z} + g ]^{T}$. \\
It can be defined that $z_{B}$ is the $z$ axis of the quadcopter in the body frame. The formula is:\\
\begin{equation}
z_{B} = \frac{t}{\left \|t\right\|} \label{zb}
\end{equation}
The intermediate coordinate system C can be used. The coordinate system C is obtained by rotating $\psi^{ref}$ rad around the z axis of the earth coordinate system. The definition formula of the x axis and the y axis of the coordinate system C are as follows:\\
\begin{equation}
x_{C} = [\cos\psi \ \sin\psi \ 0]^{T} \label{xc}
\end{equation}
\begin{equation}
y_{C} = [-\sin\psi \ \cos\psi \ 0]^{T} \label{yc}
\end{equation}
The desired thrust is :\\ 
\begin{equation}
F = m \left \|t\right\|
\end{equation}
\begin{figure}[h]
	\centering
	\includegraphics[width=8cm,height=4cm]{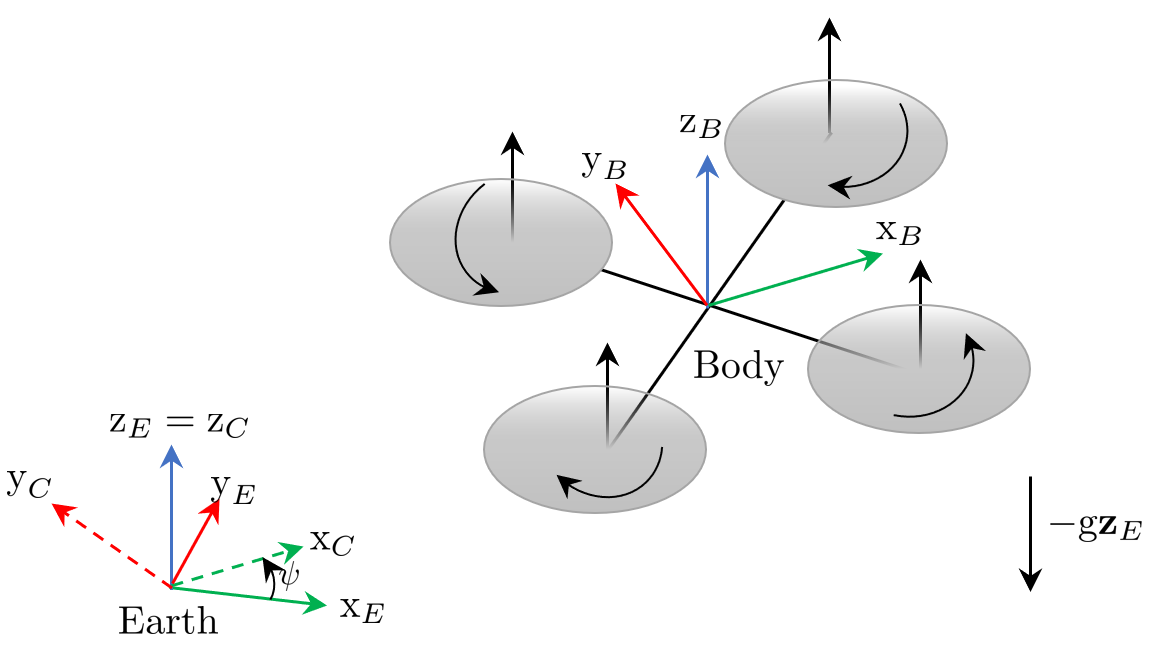}
	\caption{Quadcopter model with the used coordinate systems}\label{Quadcopter model with the used coordinate systems}
\end{figure}
The coordinate systems based on the paper by Faessler\cite{term20} are shown in figure (\ref{Quadcopter model with the used coordinate systems}). In order to align the projection of the x axis of the body coordinate system on the x-y plane of the Earth coordinate system with the x axis of the C coordinate system, the C coordinate system needs to be rotated around the z axis of the Earth coordinate system. Thus, the x axis of the body coordinate system lies on a plane composed of the x axis of the C coordinate system and the x axis of the Earth coordinate system. This is shown by:\\
\begin{equation}
x_{B} = \frac{y_{C} \times z_{B}}{\left\|y_{C} \times z_{B}\right\| } \label{xb}
\end{equation}
$z_{B}$ and $x_{B}$ are now known. The cross product of $z_{B}$ and $x_{B}$ is the direction of $y_{B}$.\\
\begin{equation}
y_{B} = \frac{z_{B} \times x_{B}}{\left\|z_{B} \times x_{B}\right\| } \label{yb}
\end{equation}
With the three axes of the body coordinate system, the rotation matrix $R^{E}_{B}$ can be obtained as follows:\\
\begin{equation}
R^{E}_{B} = [x_{B} \ y_{B} \ z_{B}]
\end{equation}\\
After the rotation matrix is obtained, the reference Euler angles $\theta^{ref}_{B} = [\phi^{ref}, \theta^{ref}, \psi^{ref}]$ can be obtained via the following formula:\\
\begin{equation}
\theta^{ref} = \text{arcsin}(- R^{E}_{B}(2,0))
\end{equation}
\begin{equation}
\phi^{ref} = \text{arctan2}(\frac{R^{E}_{B}(2,1)}{\cos\theta^{ref}}, \frac{R^{E}_{B}(2,2)}{\cos\theta^{ref}})
\end{equation}
\begin{equation}
\psi^{ref} = \text{arctan2}(\frac{R^{E}_{B}(1,0)}{\cos\theta^{ref}}, \frac{R^{E}_{B}(0,0)}{\cos\theta^{ref}})
\end{equation}
$R^{E}_{B}(i,j)$ is the element in row $i$, column $j$.\\
To calculate the desired Euler angle rotation rate, the difference between the reference Euler angel and the estimated Euler angle is needed. According to the LQR by Lucas M. Argentim \cite{term21}, $\dot{\theta}_{B}$ can be defined as:\\
\begin{equation}
\dot{\theta}_{B} = - K_{\theta} (\theta_{B} - \theta^{ref}_{B})
\end{equation}
$K_{\theta}$ is the gain matrix for the Euler angle. After many experiments, the gain matrix $K_{\theta}$ is chosen as:\\
\begin{equation}
K_{\theta}
=
\left[
\begin{array}{cccc}
40 & 0 & 0\\
0 & 40 & 0\\
0 & 0 & 40
\end{array}
\right ] \label{inertialpara}
\end{equation}
According to the relationship between Euler angle and angular velocity, the reference angular velocity $\omega^{ref}$ can be obtained from the following equation :\\
\begin{equation}
\omega^{ref}_{B} = E(\theta_{B})\dot{\theta}_{B}
\end{equation}
$E(\theta_{B})$ will be given as:\\
\begin{equation}
E(\theta_{B}) =
\left[
\begin{array}{cccc}
1 & 0 & -\sin\theta \\
0 & \cos\phi & \cos\theta\sin\phi \\
0 & -\sin\phi & \cos\theta\cos\phi
\end{array}
\right]
\end{equation}
The output of differential flatness is :\\
\begin{equation}
\text{ref} = [\text{pos}^{ref}, \text{vel}^{ref}, \theta^{ref}_{B}, \omega^{ref}_{B}]^{T}
\end{equation}

\section{Sensor} \label{sensor}
A sensor module is embedded in the quadcopter. Three different sensor schemes will be presented in this section.
\subsection{Sensor A}
It is assumed that this sensor module can observe position, acceleration and angular velocity, but output data is noisy. There are many designs of sensors. For example, noise or a linear function  can be added to the system state. In this paper, random white noise is considered. \\
Position $r$, acceleration $a_{f}$ and angular velocity $\omega$ are extracted from the quadcopter system and are sent to the sensor module.  $a_{f}$ is the acceleration caused by the actual external force and $a_{f} = [a_{fx}, a_{fy}, a_{fz}]^{T}$. $r_{n}$, $a_{n}$ and $\omega_{n}$ are the gaussian white noise of the position, acceleration and angular velocity respectively, where $r_{n} = [r_{nx}, r_{ny}, r_{nz}]^{T} $, $a_{n} = [a_{nx}, a_{ny}, a_{nz}]^{T} $ and $w_{n} = [w_{nx}, w_{ny}, w_{nz}]^{T}$.\\
Then sensor outputs position $r_{o}$, where $r_{o} = [x, y, z]$, acceleration $f$, where $f = [f_{x}, f_{y}, f_{z}]^{T}$, and angular velocity $\omega_{o}$, where $\omega_{o} = [\omega_{ox}, \omega_{oy}, \omega_{oz}]^{T}$. So they can be obtained as follows:\\
\begin{equation}
r_{o} = r + r_{n}
\end{equation}
\begin{equation}
\omega_{o} = \omega + \omega_{n}
\end{equation}
\begin{equation}
f = a_{f} + a_{n}
\end{equation}
Sensor A  has been mainly used in the quadcopter simulator. Because the sensor observes very little information, an extended Kalman filter needs to be added. The extended Kalman filter can filter out the noise and estimate the missing information, such as velocity and Euler angle. In the following parameter estimation module, if the observation of sensor A is directly used, the efficiency of parameter estimation will be reduced and the error will be very large. Therefore, the state estimation of EKF will be used for parameter estimation instead of the observation from sensor A.
\subsection{Sensor B}
Type B sensors can observe position, velocity, Euler angle, and angular velocity. The effects of observation sources on parameter estimation will be studied. Sensor B can observe all components of the state vector, so the data observed by sensor B can be directly used for parameter estimation. Sensor B can also be called a full-state observation sensor. Similar to sensor A, the output of sensor B can be defined as follows:\\
\begin{equation}
r_{o} = r + r_{n}
\end{equation}
\begin{equation}
v_{o} = v + v_{n}
\end{equation}
\begin{equation}
\theta_{o} = \theta + \theta_{n}
\end{equation}
\begin{equation}
\omega_{o} = \omega + \omega_{n}
\end{equation}
$r_{o}$, $v_{o}$, $\theta_{o}$ and $\omega_{o}$ are position, velocity, Euler angle and angular velocity respectively. Where $v_{o} = [v_{x}, v_{y}, v_{z}]^{T}$ and $\theta_{o} = [\phi, \theta, \psi]^{T}$. $r_{n}$, $v_{n}$, $\theta_{n}$ and $\omega_{n}$ are the corresponding gaussian white noise. Where $v_{n} = [v_{nx}, v_{ny}, v_{nz}]^{T} $, $\theta_{n} = [\phi_{n}, \theta_{n}, \psi_{n}]^{T} $.
\subsection{Sensor C}
In order to distinguish it from sensor B, sensor C can only observe position, velocity and angular velocity. The Euler angle can not be observed. The type C sensor is named partial observation sensor. The observations from the partial observation sensor will be stored in the database and then observations will be retrieved by the parameter estimation module. The output of sensor C can be defined as:\\
\begin{equation}
r_{o} = r + r_{n}
\end{equation}
\begin{equation}
v_{o} = v + v_{n}
\end{equation}
\begin{equation}
\omega_{o} = \omega + \omega_{n}
\end{equation}
Similar to sensor A and sensor B, $r_{o}$, $v_{o}$ and $\omega_{o}$ are position, velocity and angular velocity respectively. $r_{n}$, $v_{n}$ and $\omega_{n}$ are the corresponding gaussian white noise.
\section{State space model for extended Kalman filter}
The extended Kalman filter is used to estimate the multidimensional state. The state includes position, velocity, Euler angle, and angular velocity. The following nonlinear system equations describe the relation between states at adjacent moments, as well as the relation between observations and system states.\\
\begin{equation}
x_{k} = f(x_{k-1},u_{k-1},n_{k}) , \ n_{k} \sim N(0,Q)
\end{equation}
$n_{k}$ is process noise and obeys Gaussian distribution. $Q$ is the covariance of the process noise.\\
\begin{equation}
y_{k} = h(x_{k},v_{k}) , \ v_{k} \sim N(0,R)
\end{equation}
$ v_{k}$ is observation noise and obeys Gaussian distribution. $R$ is the covariance of the observation noise.\\
State variables are given in the form of vectors. The state is given as follows:\\
\begin{equation}
X_{k} = [x, y ,z , v_{x}, v_{y}, v_{z}, \phi, \theta, \psi, \omega_{x}, \omega_{y}, \omega_{z}]^{T}
\end{equation}
In this state vector, position is $r = [x,y,z]^{T}$, velocity is $v = [v_{x}, v_{y}, v_{z}]^T$, $\theta_{B} = [\phi, \theta, \psi]^{T}$ is attitude angle. $\omega_{B} = [\omega_{x}, \omega_{y}, \omega_{z}]^{T}$ is angular velocity.
\section{Expectation maximization algorithm}
Expectation maximization algorithm is a commonly used algorithm in parameter estimation problems. The observation $Y$ is known, but $X$ is unknown.  $X$ and $Y$ can be obtained from the joint probability distribution $p(x,y|\theta)$. And the joint probability distribution $p(x,y|\theta)$ depends on parameter $\theta$. The goal is to estimate the model parameters by using the EM algorithm and the observation data $Y$. Based on the maximum likelihood estimate, it is defined as:\\
\begin{equation}
\begin{split}
\hat{\theta} &= \mathop{\arg\max}_{\theta} \ p(y|\theta)\\
& = \mathop{\arg\max}_{\theta} \ \log p(y|\theta)
\end{split}
\end{equation}
Because $p(y|\theta )$ contains an integral over an unknown variable $X$, it is difficult to find the optimal solution of $\theta$. Optimization problems can be simplifed. An auxiliary distribution $q$ and Jesens's inequality can be applied to move the integral outside of the logarithm.\\
\begin{equation}
\begin{split}
\log p(y|\theta) &= \log \left(\int p(x, y|\theta) dx \right) \\
&= \log \left(\int q(x)\frac{p(x,y|\theta)}{q(x)}dx\right) \\
&\geq \int q(x) \log \left(\frac{p(x,y|\theta)}{q(x)}\right) dx 
\end{split}
\end{equation}
So it results in:\\
\begin{equation}
L(q,\theta) = \int q(x) \log \left(\frac{p(x,y|\theta)}{q(x)}\right) dx 
\end{equation}
$L(q,\theta)$ is a lower bound for $\log p(y|\theta)$ which contains the unknown X. Trying to maximize this lower bound equivalents to trying to maximize $\log p(y|\theta)$. So $L(q,\theta)$ needs to be maximized. The EM algorithm has E and M steps. First an initial value for $\theta^{0}$ has been chosen. For $i = 1,2,...n$  \\
E step:\\
\begin{equation}
q^{(i)} = \mathop{\arg\max}_{q} \ L(q,\theta^{(i-1)})
\end{equation}
M step:\\
\begin{equation}
\theta^{(i)} = \mathop{\arg\max}_{\theta} \ L(q^{(i)},\theta)
\end{equation}
The above two formulas describe that in E step, it updates $q^{(i)}$ base on current parameter $\theta^{(i-1)}$. Then, the $q^{(i)}$ from E step is put into M step to update the parameter $\theta^{(i)}$. The above steps are repeated until parameter $\theta$ converges.
First the q update of E step will be analyzed. Expanding the joint probability, it results in:\\
\begin{equation}
p(x,y|\theta) = p(x|y,\theta)p(y|\theta)
\end{equation}
The objective function can be simplified as:\\
\begin{equation}
\begin{split}
L(q,\theta^{(i-1)}) &= 
\int q(x) \log \Bigg(\frac{p(x \mid y, \theta^{(i-1)})}{q(x)} \Bigg) dx \\
&\quad + \int q(x) \log p(y \mid \theta^{(i-1)}) dx \\
&= 
\int q(x) \log \Bigg(\frac{p(x \mid y, \theta^{(i-1)})}{q(x)} \Bigg) dx 
+ \log p(y \mid \theta^{(i-1)})
\end{split}
\end{equation}

Removing terms that are not related to $q$, it results in:\\
\begin{equation}
\begin{split}
q^{(i)} &= \mathop{\arg\max}_{q} \left\{\int q(x) \log \left(\frac{p(x|y,\theta^{(i-1)})}{q(x)}\right)dx + \text{const}\right\}\\
&= \mathop{\arg\min}_{q} \ \left\{ D_{KL}[q(x)||p(x|y,\theta^{i-1})]\right\} \\
&= p(x|y,\theta^{(i-1)})
\end{split}
\end{equation}
In the above equation,$D_{KL}$ means (Kullback–Leibler) divergence\cite{term18} which is used to distinguish the difference between one distribution and another. The more similar the two distributions, the smaller the $KL$ divergence, and the minimum value is 0, which means that the two distributions are the same. So this means that the goal of E step is to find the accurate hidden variable $X$. Next the M step will be analyzed. $q^{(i)}$ from E step will be used in M step. The function $L(q^{(i)},\theta)$ is given as follows:\\
\begin{equation}
\begin{split}
L(q^{(i)},\theta) &= \int p(x|y,\theta^{(i-1)}) \log \left(\frac{p(x,y|\theta)}{p(x|y,\theta^{(i-1)})}\right)dx \\
& =\int p(x|y,\theta^{(i-1)}) \log (p(x,y|\theta))dx +\text{const} \\ 
\end{split}
\end{equation}
The updating expression of $\theta$ can be simplified to:\\
\begin{equation}
\theta^{(i)} = \mathop{\arg\max}_{\theta} \left\{\int p(x|y,\theta^{(i-1)}) \log (p(x,y|\theta))dx \right\}
\end{equation}
In E step, the conditional probability expectations of the joint distribution will be calculated, and then the objective function with respect to parameter $\theta$ in M step will be maximized. EM algorithm is often used to learn the parameters of Gaussian mixture model (GMM), hidden Markov algorithm (HMM) and variational inference of latent Dirichlet allocation (LDA) topic model, etc. The details of the EM algorithm depend on the selected model.
\section{Dynamics} \label{dynamic}
\subsection{Dynamical model}\label{Dynamical Model}
The state space model is applied to the quadcopter system. $X$ denotes the system state. $X = [x, y, z, v_{x}, v_{y}, v_{z}, \phi, \theta, \psi, \omega_{x}, \omega_{y}, \omega_{z} ]^{T}$. Position is $r = [x, y, z]^{T}$. Velocity is $v = [v_{x}, v_{y}, v_{z}]^{T}$. In the body frame, the roll, pitch and yaw angles are defined as $\theta_{B} = [\phi, \theta, \psi]^{T}$. The corresponding roll, pitch and yaw rotation rates are $\dot{\theta}_{B} = [\dot{\phi}, \dot{\theta}, \dot{\psi}]^{T}$. Angular velocity is $\omega_{B} = [\omega_{x}, \omega_{y}, \omega_{z}]^{T}$. \\
In the system, there are two coordinates. One is the earth frame coordinate and the corresponding orthogonal basis vectors are: $[x_{E}, y_{E}, z_{E}]$ . Another one is the body frame coordinate and the corresponding orthogonal basis vectors are:$[x_{B}, y_{B}, z_{B}]$. One way to convert parameters of the body frame coordinate to the Earth frame coordinate : $x = R^{E}_{B}(\theta_{B})x_{B}$. $R(\theta_{B})$ is a rotation matrix. The rotation matrix can be obtained from the Euler angles. The formula is as follows:\\
\begin{equation}
\begin{split}
R_{B}^{E} &= R_{x}(\phi)\,R_{y}(\theta)\,R_{z}(\psi) \\
&=
\left[
\begin{array}{ccc}
c_\theta c_\psi & c_\theta s_\psi & -s_\theta \\
s_\phi s_\theta c_\psi - c_\phi s_\psi &
s_\phi s_\theta s_\psi + c_\phi c_\psi &
s_\phi c_\theta \\
c_\phi s_\theta c_\psi + s_\phi s_\psi &
c_\phi s_\theta s_\psi - s_\phi c_\psi &
c_\phi c_\theta
\end{array}
\right]
\end{split}
\end{equation}
With 
$c_\alpha = \cos\alpha, \qquad s_\alpha = \sin\alpha.
$

The inverse Euler angle rates matrix can convert the angular velocity $\omega_{B}$ to the rotation rate $\dot{\theta}_{B}$. The inverse Euler angle rates matrix is :\\
\begin{equation}
E^{-1}(\theta_{B}) =
\left[
\begin{array}{cccc}
1 & \sin\phi \tan\theta & \cos\phi \tan\theta \\
0 & \cos\phi & -\sin\phi \\
0 & \frac{\sin\phi}{\cos\theta} & \frac{\cos\phi}{\cos\theta}
\end{array}
\right]
\end{equation}\label{EthetaB}
In the system of this paper, the forces acting on the quadcopter are gravity and the forces of the propellers. According to Newton's Euler equation, it results in:\\
\begin{equation}
m\ddot{r} =
\left[
\begin{array}{cccc}
0\\
0\\
-mg
\end{array}
\right]
+ R_{B}^{E}(\theta_{B} )
\left[
\begin{array}{cccc}
0\\
0\\
F
\end{array}
\right]
\end{equation}
$m$ is the mass of the quadcopter. $r$ is the quadcopter position. $g$ is the gravitational constant. $F$ is the thrust force in the body frame coordinate. Rotating propellers generate torques. The torque $M$ can be calculated from Euler's equation. This is shown in the following formula:\\
\begin{equation}
M = I\dot{\omega}_{B} + \omega_{B}\times (I\omega_{B})
\end{equation}
$I$ is the inertia matrix.\\
\begin{equation}
I =
\left[
\begin{array}{ccc}
I_{xx} & 0 & 0\\
0 & I_{yy} & 0\\
0 & 0 & I_{zz}
\end{array}
\right]
\label{inertia}
\end{equation} 
Position is $x_{1} = r = [ x , y , z ]^{T}$. Velocity is $x_{2} = [v_{x} ,v_{y} ,v_{z}]^{T}$. Angular velocity is $x_{4} = \omega_{B} = [\omega_{x},\omega_{y},\omega_{z}]^{T}$. The following controlled dynamical system is the ODE model. Control vector is $u = [F,M]^{T}$. Thrust is $F$. Torque $M = [M_{x},M_{y},M_{z}]^{T}$. \\
\begin{equation}
\begin{aligned}
\dot{x} = f(x,u)
\end{aligned}
\end{equation}
The ODE nonlinear dynamical system based on the paper by Andrew\cite{term0} is given as follows:\\
\begin{equation}
\left[
\begin{array}{cccc}
\dot{x}_{1}\\
\dot{x}_{2}\\
\dot{x}_{3}\\
\dot{x}_{4}
\end{array}
\right ]
=
\left[
\begin{array}{cccc}
x_{2}\\
-g\mathbf{e_{3}} + \frac{1}{m}R_{B}^{E}(x_{3})F\\
E^{-1}(x_{3})x_{4} \\
I^{-1}(M - \omega_{B}\times{I\omega_{B}} )
\end{array}
\right ]
\end{equation}
$g$ is gravity. $\mathbf{e_{3}}$ is $[0,0,1]^{T}$. $R$ is the rotation matrix. \\
In practice, the force acting on a quadcopter is uncertain. Because there are many uncertain or unknown factors around the quadcopter, such as wind and inaccurate measurement instruments. All of these uncertain factors have a non negligible impact on quadcopter systems. So next, random noise is added to the system model to change the ODE model to the SDE model. There is a stochastic differential equation form:\\
\begin{equation}
\begin{aligned}
\text{dx} = f(x,u)dt + Gdw
\end{aligned}\label{sdeevo}
\end{equation}
Random noise is added to the control vector. In other words, random noise is added to thrust $F$ and torque $M$. The SDE nonlinear dynamical system is given as follows:\\
\begin{equation}
\left[
\begin{array}{cccc}
\dot{x_{1}}\\
\dot{x_{2}}\\
\dot{x_{3}}\\
\dot{x_{4}}
\end{array}
\right ]
=
\left[
\begin{array}{cccc}
x_{2}\\
-g\mathbf{e_{3}} + \frac{1}{m}R_{B}^{E}(x_{3})(F+n_{0})\\
E^{-1}(x_{3})x_{4} \\
I^{-1}((M+n_{1:3}) - \omega_{B}\times{I\omega_{B}} )
\end{array}
\right ],
\end{equation}
Where $ n\sim N(0,hI)$, $n$ is gaussian white noise and $hI$ is the covariance of the process noise.\\
Let this system evolution of the state model adapt to the form of equation (\ref{sdeevo}). The SDE equation can be obtained as follows: \\
\begin{equation}
\begin{aligned}
\left[
\begin{array}{cccc}
\dot{x_{1}}\\
\dot{x_{2}}\\
\dot{x_{3}}\\
\dot{x_{4}}
\end{array}
\right ]
&=
\underbrace{
	\left[
	\begin{array}{cccc}
	x_{2}\\
	-g\mathbf{e_{3}} + \frac{1}{m}R_{B}^{E}(x_{3})F\\
	E^{-1}(x_{3})x_{4} \\
	I^{-1}(M - \omega_{B}\times{I\omega_{B}} )
	\end{array}
	\right ]}_{f(x,u)}
\\[6pt]
& +
\underbrace{
	\left[
	\begin{array}{cccc}
	0 & 0\\
	\frac{1}{m}R_{B}^{E}(x_{3})  & 0 \\
	0 & 0\\
	0 & I^{-1}
	\end{array}
	\right ]}_{G}
\left[
\begin{array}{cccc}
n_{0} \\
n_{1:3}
\end{array}
\right ]
\end{aligned}
\end{equation}
$n\sim N(0,hI)$.\\
The current state of the system depends only on the previous state. So the hidden Markov chain can be used to update the state. For a small step $dt$, it results in:\\
\begin{equation}
\begin{aligned}
X_{k} = X_{k-1} + \text{dx} =  X_{k-1} + f(x,u)dt + Gdw 
\end{aligned}
\end{equation}
The measurement model is the same as equation (\ref{state2}) mentioned above.
\subsection{Parameters for the quadcopter system}
The key parameters for the quadcopter are:\\
(a) Mass: $m$ = 0.18 kg;\\
(b) Gravity: $g$ = 9.81 $\text{m/s}^{2}$;\\
(c) The distance between the center of mass and the axis of motor: L = 0.086m ;\\
(d) The inertia matrix $I$ is shown below. The unit of the inertia matrix $I$ is $\text{kg} \cdot \text{m}^{2}$.\\
\begin{equation}
I
=
\left[
\begin{array}{cccc}
2.5\text{E-}4 & 0 & 0\\
0 & 3.1\text{E-}4 & 0\\
0 & 0 & 2\text{E-}4
\end{array}
\right ] \label{inertialpara}
\end{equation}
\section{State estimation}
The quadcopter can acquire its own position,  acceleration and angular velocity by using sensor A. But these are noisy data, which will seriously affect the flight of the quadcopter. So an extended Kalman filter has been used to filter out state noise and restore the missing information.\\
Kalman filters are suitable for linear systems. The observation data is used in the Kalman system, and then the state of the system is optimally estimated. Optimal estimation is a filtering process. An extended Kalman filter has been used in this paper as the extended Kalman filter is suitable for nonlinear systems. The extended Kalman filter linearizes a non-linear system and then performs a Kalman filter. The extended Kalman filter retains the first-order terms of the Taylor expansion of the nonlinear function and ignores the remaining higher-order terms.
\subsection{Prediction model}
The dymanic model is discrete and the dynamic equation of a non-linear quadcopter is shown below :\\
\begin{equation}
\left[
\begin{array}{cccc}
r\\
v\\
\theta_{B}\\
\omega_{B}
\end{array}
\right ]_{k}
= 
\left[
\begin{array}{c}
r\\
v\\
\theta_{B}\\
\omega_{B}
\end{array}
\right]_{k-1}
+ \triangle t 
\left[
\begin{array}{cccc}
v\\
f\\
E^{-1}(\theta_{B})\omega_{B} \\
0 
\end{array}
\right ] + n_{k},
\label{prediction}
\end{equation}
 $\ n_{k} \sim N(0,Q)$, $n_{k}$ is system noise. From the above dynamic equation the Jacobian matrix $\Phi$ can be derived as follows:\\
\begin{equation}
\Phi_{k-1} =
\resizebox{\columnwidth}{!}{$
\left[
\begin{array}{cccccccccccc}
1 & 0 & 0 & \Delta t & 0 & 0 & 0 & 0 & 0 & 0 & 0 & 0 \\
0 & 1 & 0 & 0 & \Delta t & 0 & 0 & 0 & 0 & 0 & 0 & 0 \\
0 & 0 & 1 & 0 & 0 & \Delta t & 0 & 0 & 0 & 0 & 0 & 0 \\
0 & 0 & 0 & 1 & 0 & 0 & 0 & 0 & 0 & 0 & 0 & 0 \\
0 & 0 & 0 & 0 & 1 & 0 & 0 & 0 & 0 & 0 & 0 & 0 \\
0 & 0 & 0 & 0 & 0 & 1 & 0 & 0 & 0 & 0 & 0 & 0 \\
0 & 0 & 0 & 0 & 0 & 0 & \frac{\partial\phi}{\partial\phi} &
\frac{\partial\phi}{\partial\theta} & 0 & \Delta t &
\frac{\partial\phi}{\partial\omega_{y}} &
\frac{\partial\phi}{\partial\omega_{z}} \\
0 & 0 & 0 & 0 & 0 & 0 & \frac{\partial\theta}{\partial\phi} & 1 & 0 & 0 &
\frac{\partial\theta}{\partial\omega_{y}} &
\frac{\partial\theta}{\partial\omega_{z}} \\
0 & 0 & 0 & 0 & 0 & 0 & \frac{\partial\psi}{\partial\phi} &
\frac{\partial\psi}{\partial\theta} & 1 & 0 &
\frac{\partial\psi}{\partial\omega_{y}} &
\frac{\partial\psi}{\partial\omega_{z}} \\
0 & 0 & 0 & 0 & 0 & 0 & 0 & 0 & 0 & 1 & 0 & 0 \\
0 & 0 & 0 & 0 & 0 & 0 & 0 & 0 & 0 & 0 & 1 & 0 \\
0 & 0 & 0 & 0 & 0 & 0 & 0 & 0 & 0 & 0 & 0 & 1
\end{array}
\right]
$}
\label{jacobianF}
\end{equation}

$f = [f_{x}, f_{y}, f_{z}]^T$ and $\omega = [\omega_{x}, \omega_{y}, \omega_{z}]^T$ are given by the sensor. Since some terms of the Jacobian matrix $\Phi$ are very long, the unexpanded terms will be explained below.\\
\begin{equation}
\frac{\partial\phi}{\partial\phi} =1 +\triangle t( \cos\phi \tan\theta \omega_{y} - \sin\phi \tan\theta \omega_{z}) 
\end{equation}
\begin{equation}
\frac{\partial\phi}{\partial\theta} =  \triangle t (\sin\phi \sec^{2}(\theta) \omega_{y} + \cos\phi \sec^{2}( \theta) \omega_{z}) 
\end{equation}
\begin{equation}
\frac{\partial\phi}{\partial\omega_{y}} = \triangle t \sin\phi \tan\theta 
\end{equation}
\begin{equation}
\frac{\partial\phi}{\partial\omega_{z}} = \triangle t \cos\phi \tan\theta 
\end{equation}
\begin{equation}
\frac{\partial\theta}{\partial\phi} = \triangle t( -\sin\phi \omega_{y} - \cos\phi \omega_{z}) 
\end{equation}
\begin{equation}
\frac{\partial\theta}{\partial\omega_{y}} = \triangle t \cos\phi 
\end{equation}
\begin{equation}
\frac{\partial\theta}{\partial\omega_{z}} =  -\sin\phi \triangle t 
\end{equation}
\begin{equation}
\frac{\partial\psi}{\partial\phi} = \triangle t( \frac{\cos\phi}{\cos\theta} \omega_{y} - \frac{\sin\phi}{\cos\theta} \omega_{z} )
\end{equation}
\begin{equation}
\frac{\partial\psi}{\partial\theta} = \triangle t( \sin\phi \frac{\sin\theta}{\cos^{2}\theta} \omega_{y} + \cos\phi \frac{\sin\theta}{\cos^{2}\theta} \omega_{z} )
\end{equation}
\begin{equation}
\frac{\partial\psi}{\partial\omega_{y}} = \triangle t \frac{\sin\phi}{\cos\theta} 
\end{equation}
\begin{equation}
\frac{\partial\psi}{\partial\omega_{z}} = \triangle t \frac{\cos\phi}{\cos\theta} 
\end{equation}
The formula (\ref{prediction}) is the prediction step of the extended Kalman filter. This step uses the estimated value of the previous step.
\subsection{Measurement model}
The measurement model outputs noise observations. As explained in Section (\ref{sensor}), the sensor A can observe position, acceleration and angular velocity. The observation equation of the extended Kalman filter is given as follows:\\
\begin{equation}
y_{k} = H_{k}x_{k} + v_{k}, \ v_{k} \sim N(0,R)
\end{equation}
$v_{k}$ is measurement noise. $H_{k}$ is the observation matrix and is given as follows :\\
\begin{equation}
H_{k} = \text{diag}[1, 1, 1, 0, 0, 0, 0, 0, 0, 1, 1, 1]
\end{equation}
\subsection{Correction}
In this step, the Kalman gain matrix will be introduced. Whether the estimated state depends more on the predicted state or the measured state will be weighed by the gain matrix. In order to facilitate the distinction, symbol \ $\hat{}$ \ has been used to represent the estimated value and symbol \ $\check{}$ \ has been used to represent the predicted value. The Kalman gain matrix $K_{k}$ is given by:\\
\begin{equation}
K_{k} = \check{P}_{k} H^{T}_{k}(H_{k}\check{P}_{k}H^{T}_{k} + R_{k})^{-1}
\end{equation}
In the above formula, $\check{P}$ is the a priori estimated covariance at time $k$, $\hat{P}_{0}$ will be given as the initial condition, and covariance $P$ will converge quickly. The recursive formula for covariance $P$ is:\\
\begin{equation}
\check{P}_{k} = \hat{\Phi}_{k-1}\hat{P}_{k-1}\hat{\Phi}^{T}_{k-1} + Q_{k} 
\end{equation}
Covariance of the posterior estimate at time $k$ is:\\
\begin{equation}
\hat{P}_{k} = \check{P}_{k} - K_{k}H_{k} \check{P}_{k}
\end{equation}
$\Phi$ is the Jacobian matrix of the prediction model, and $Q$ is the covariance of the process noise.\\
Next, the estimated state will be updated. In the updating step, the Kalman gain is multiplied by the difference between the measured state $y_{k}$ and the predicted state $\check{x}_{k}$. The estimated state is given as:\\
\begin{equation}
\hat{x}_{k} = \check{x}_{k} + K_{k}[y_{k} - H_{k}\check{x}_{k}]
\end{equation}
\section{Extended Kalman filter results}
\begin{figure}[h]
	\centering
	\includegraphics[width=\columnwidth]{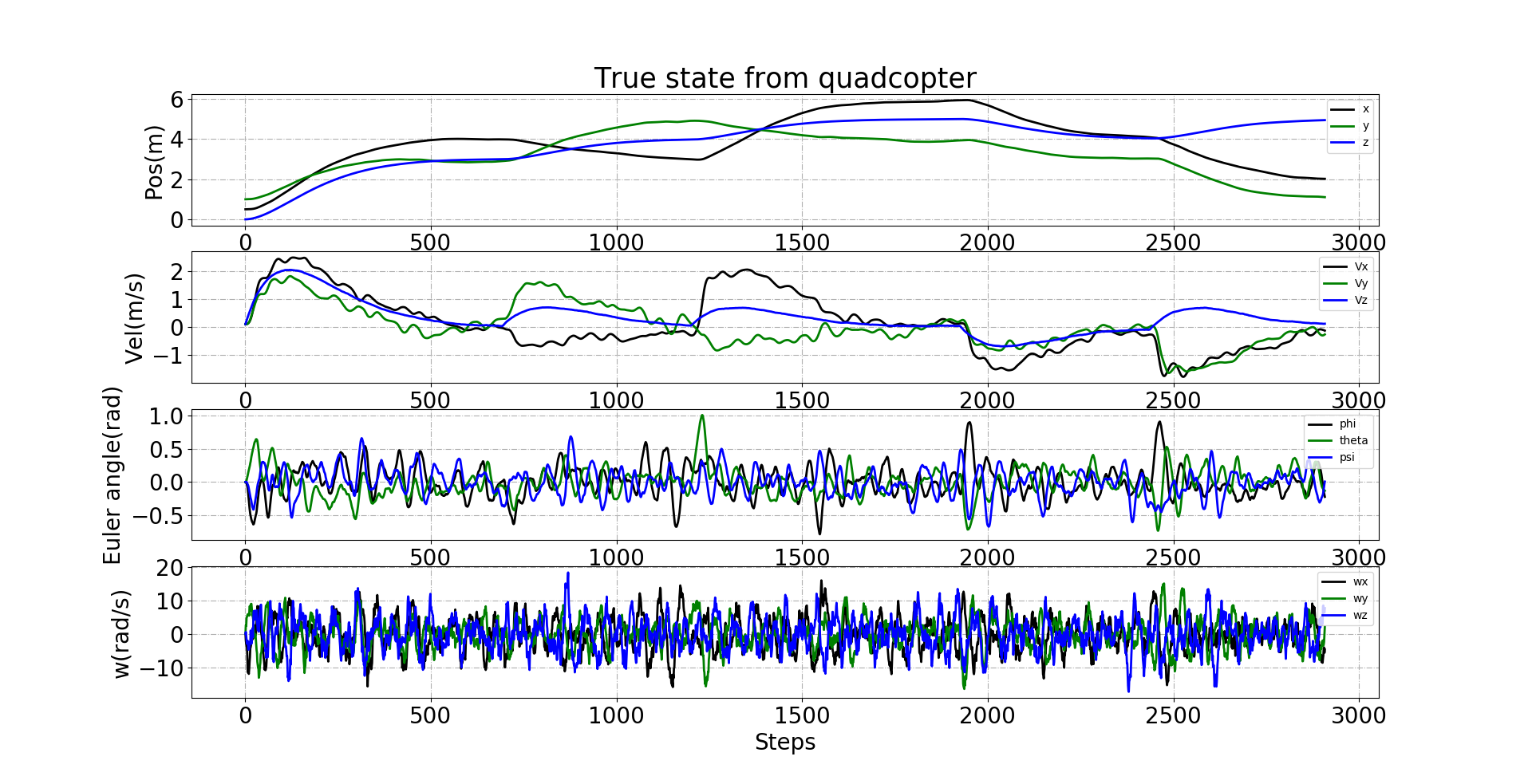}
	\caption{True state from quadcopter} \label{True state from quadcopter}
\end{figure}

 In this section, the results of the extended Kalman filter will be presented. The extended Kalman filter plays an important role in the quadcopter simulator in this paper.\\
 Figure (\ref{True state from quadcopter}) shows the true state of the quadcopter without measurement noise, which is the true trajectory of the quadcopter. This picture includes four parts. The first part is the trajectory of position $[x, y, z]^T$, which is represented by black, green and blue lines respectively. The second part is the trajectory of the speed $[v_{x}, v_{y}, v_{z}]^T$, which also uses black, green and blue lines to represent $v_{x}, v_{y}, v_{z}$. The third part is the trajectory of the Euler angle $[\phi, \theta, \psi]^T$, which is represented by black, green and blue lines respectively. The fourth part is the trajectory of angular velocity $[\omega_{x}, \omega_{y}, \omega_{z}]^T$, which is also represented by black, green and blue lines.
\begin{figure}[t]
	\centering
	\includegraphics[width=\columnwidth]{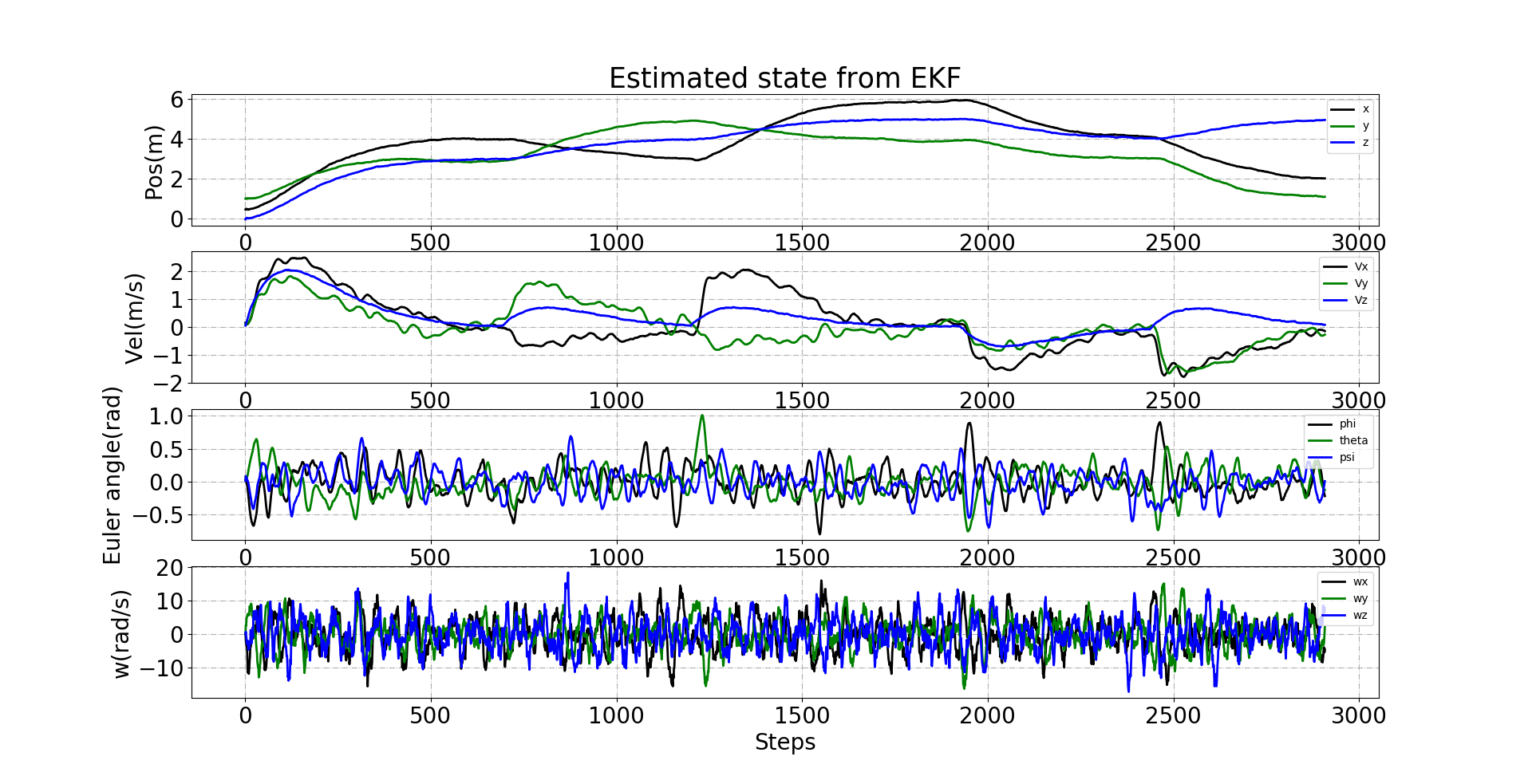}
	\caption{Estimated state from EKF} \label{Estimated state from EKF}
\end{figure}
Figure (\ref{Estimated state from EKF}) is the state estimation result by using the extended Kalman filter. The first part is the trajectory of position $[x, y, z]^T$, which uses black, green and blue lines to represent $x$, $y$, $z$. The second part is the trajectory of the velocity $[v_{x}, v_{y}, v_{z}]^T$, which uses black line, green line and blue line to represent $v_{x}$, $v_{y}$, $v_{z}$. The third part is the trajectory of the Euler angle $[\phi, \theta, \psi]^T$, which is represented by black, green and blue lines respectively. The fourth part is the trajectory of angular velocity $[\omega_{x}, \omega_{y}, \omega_{z}]^T$, which is represented by black, green and blue lines. Because the sensor can only observe acceleration and angular acceleration, the EKF is used to restore the state based on acceleration and angular acceleration.
\begin{figure}[t]
	\centering
	\includegraphics[width=\columnwidth]{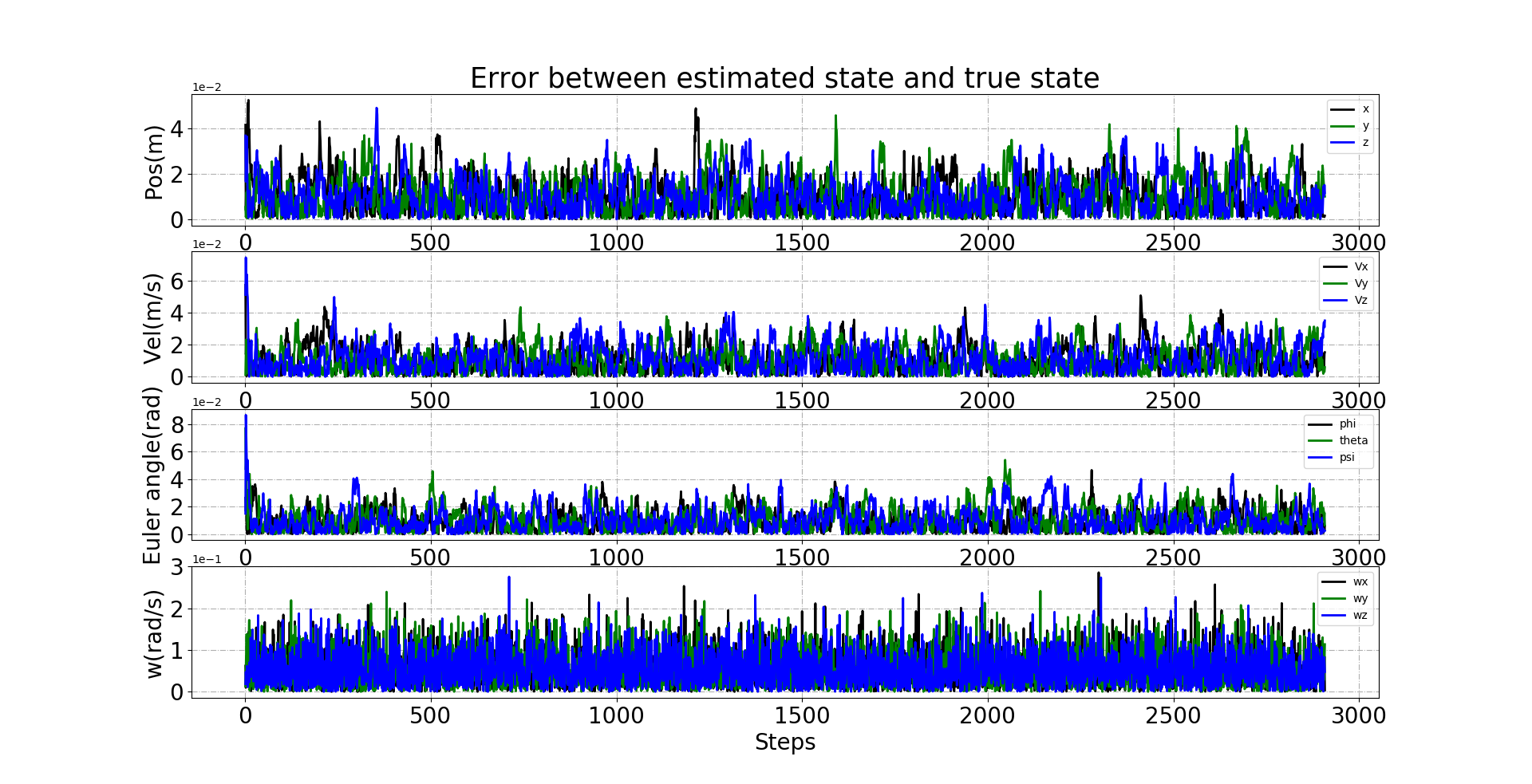}
	\caption{Error between eatimated state and true state} \label{Error between eatimated state and true state_known}
\end{figure}
Figure (\ref{Error between eatimated state and true state_known}) shows the error between the results of the extended Kalman filter and the true state of the quadcopter. The first part of figure (\ref{Error between eatimated state and true state_known}) is position error, the second part is velocity error, the third part is Euler angle error, and the last part is angular velocity error. The position error is below 3E-2 m for almost the whole trajectory. Velocity error is controlled below 3E-2 m/s for almost the whole trajectory. The Euler angle error is controlled below 3E-2 rad for almost the whole trajectory. For the angular velocity, it can be seen from figure (\ref{Error between eatimated state and true state_known}) that the error is controlled between 0 and 1.5E-1 rad/s for most of the time.
\begin{table}[h]
	\renewcommand\arraystretch{1.5}
	\centering
	\setlength{\abovecaptionskip}{0pt}%
	\setlength{\belowcaptionskip}{10pt}%
	\caption{Comparison of pose estimation performance}
	\begin{tabular}{c|c|c|c}
		\hline  
		& \cite{term13} & \cite{term14}  & Our system \\ 
		\hline  
		Mean position error in m	  &  1.41 E-2  &  1.5 E-2	& 1.02 E-2  \\
		
		Standard deviation in m &  1.02 E-2 & 0.7 E-2 & 0.77 E-2  \\
		
		Max position error in m &  11.2 E-2 & 12.1 E-2 & 5.24 E-2  \\
		\hline 
		Mean Euler angle error in rad & 2.67 E-2  &  2.09 E-2 & 1.04 E-2 \\
		Standard deviation in rad & 2.81 E-2 & 0.7 E-2 & 0.83 E-2  \\
		
		Max Euler angle error in rad	 & 34 E-2 & 7.85 E-2 & 8.67 E-2  \\
		\hline 
	\end{tabular}\label{Comparison of pose estimation performance}
\end{table}
In table (\ref{Comparison of pose estimation performance}), the results of the performance are compared with those in the paper by Faessler\cite{term10}. In the paper by Faessler\cite{term10}, the results of pose estimation performance are based on the paper by Edwin\cite{term13} and Breitenmoser\cite{term14}. The mean position error of EKF in this theis is 29\% smaller than the mean position error in the paper by Edwin\cite{term13}  and 33.3\% smaller than the result in \cite{term14}. The standard deviation of position error in our system is 24.5 \% lower than in\cite{term13} and 9 \% higher than in\cite{term14}. The max position error of our system is smaller than the results in\cite{term13} and \cite{term14}.\\
The mean Euler angle error of our system is 61\% smaller than the mean Euler angle error in the paper by Edwin\cite{term13}  and 50\% smaller than the result in \cite{term14}. The standard deviation of Euler angle error in our system is 70 \% lower than in \cite{term13} and 15.7 \% higher than in \cite{term14}. The max Euler angle error of our system is 74.5\% smaller than the results in \cite{term13} and 9.46\% higher than in \cite{term14}.\\
Therefore, the filtering effect of the extended Kalman filter is good in comparison to the above mentioned data from \cite{term10}. It can filter out the noise and try to restore the true state from the noise interference information.
\section{LQG controller}
LQG is short for linear quadratic Gaussian control. Differential flatness can help to calculate the feed forward input from desired trajectory. LQG controller enables the quadcopter to track trajectories or fly to the target point.\\
From the previous Section, it is alrealdy known that the extended Kalman filter can remove process noise and measurement noise of the state. The results of the extended Kalman filter have been sent to the LQG controller.\\
State has been defined as: \\

$ \hat{X} = [ x,y,z, v_{x}, v_{y}, v_{z}, \phi, \theta, \psi, \omega_{x}, \omega_{y}, \omega_{z}]^T $\\
Reference state has been defined as: \\
$X^{ref} = [ x^{ref}, y^{ref}, z^{ref}, v_{x}^{ref}, v_{y}^{ref},  v_{z}^{ref}, \phi^{ref}, \theta^{ref}, \psi^{ref},\\ \omega_{x}^{ref}, \omega_{y}^{ref}, \omega_{z}^{ref}]^T$ \\
When the Euler angles are all small angles, the following approximation can be obtained:\\
\begin{equation}
\left[
\begin{array}{ccc}
\omega_{x} \\
\omega_{y} \\
\omega_{z}
\end{array}
\right] \approx
\left[
\begin{array}{ccc}
\dot{\phi} \\
\dot{\theta}\\
\dot{\psi}
\end{array}
\right]
\end{equation}
The following state space representation can be obtained as follows:\\ 
\begin{equation}
\dot{X} = f(X,u) = 
\left\{
\begin{array}{lr}
\dot{x}_{1} = x_{4} \\
\dot{x}_{2} = x_{5} \\
\dot{x}_{3} = x_{6} \\
\dot{x}_{4} = ua_{x} \\
\dot{x}_{5} = ua_{y} \\
\dot{x}_{6} = ua_{z} \\
\dot{x}_{7} = x_{10} \\
\dot{x}_{8} = x_{11} \\
\dot{x}_{9} = x_{12} \\
\dot{x}_{10} = \dot{\omega}_{x} \\
\dot{x}_{11} = \dot{\omega}_{y} \\
\dot{x}_{12} = \dot{\omega}_{z}  
\end{array}
\right.
\end{equation}
The discretized linear model of the LQG controller is given below:\\
\begin{equation}
\dot{x}_{k} = \mathbf{A}x_{k} + \mathbf{B}u_{k} 
\end{equation}
The A jacobian matrix and B jacobian matrix can be obtained from following expression:\\ 
\begin{equation}
\mathbf{A}
=
\left[
\begin{array}{cccccccccccc}
\frac{\partial{f_{1}}}{\partial x_{1} } & \frac{\partial{f_{1}}}{\partial x_{2} } & \cdots & \frac{\partial{f_{1}}}{\partial{x_{11}}} & \frac{\partial{f_{1}}}{\partial{x_{12}}} \\
\frac{\partial{f_{2}}}{\partial x_{1} } & \frac{\partial{f_{2}}}{\partial x_{2} } &  \cdots & \frac{\partial{f_{2}}}{\partial{x_{11}}} & \frac{\partial{f_{2}}}{\partial{x_{12}}} \\
\vdots & \vdots & \ddots &  \vdots  &  \vdots\\
\frac{\partial{f_{11}}}{\partial x_{1} } & \frac{\partial{f_{11}}}{\partial x_{2} } &  \cdots & \frac{\partial{f_{11}}}{\partial{x_{11}}} & \frac{\partial{f_{11}}}{\partial{x_{12}}} \\
\frac{\partial{f_{12}}}{\partial x_{1} } & \frac{\partial{f_{12}}}{\partial x_{2} } &  \cdots & \frac{\partial{f_{12}}}{\partial{x_{11}}} & \frac{\partial{f_{12}}}{\partial{x_{12}}} \\
\end{array}
\right ]
\end{equation}
\begin{equation}
\mathbf{B}
=
\left[
\begin{array}{cccccc}
\frac{\partial{f_{1}}}{\partial ua_{x} } & \frac{\partial{f_{1}}}{\partial ua_{y} } & \cdots & \frac{\partial{f_{1}}}{\partial{\dot{\omega}_{y}}} & \frac{\partial{f_{1}}}{\partial{\dot{\omega}_{z}}} \\
\frac{\partial{f_{2}}}{\partial ua_{x} } & \frac{\partial{f_{2}}}{\partial ua_{y} } & \cdots & \frac{\partial{f_{2}}}{\partial{\dot{\omega}_{y}}} & \frac{\partial{f_{2}}}{\partial{\dot{\omega}_{z}}} \\
\vdots & \vdots & \ddots &  \vdots  &  \vdots  \\
\frac{\partial{f_{11}}}{\partial ua_{x} } & \frac{\partial{f_{11}}}{\partial ua_{y} } & \cdots & \frac{\partial{f_{11}}}{\partial{\dot{\omega}_{y}}} & \frac{\partial{f_{11}}}{\partial{\dot{\omega}_{z}}} \\
\frac{\partial{f_{12}}}{\partial ua_{x} } & \frac{\partial{f_{12}}}{\partial ua_{y}} & \cdots & \frac{\partial{f_{12}}}{\partial{\dot{\omega}_{y}}} & \frac{\partial{f_{12}}}{\partial{\dot{\omega}_{z}}}
\end{array}
\right ]
\end{equation}

The $lqr$ function of Matlab can be used to calculate the gain matrix $K$.
In order to simplify the gain matrix $K$, our control $u$ is:
\begin{equation}
\begin{split}
u = -K (\hat{X} - X^{ref}) = 
\left[
\begin{array}{ccc}
ua_{x}\\
ua_{y}\\
ua_{z}\\
\dot{\omega}_{x}\\
\dot{\omega}_{y}\\
\dot{\omega}_{z}
\end{array}
\right] 
\end{split}
\end{equation}
According to the information in Section (\ref{Dynamical Model}) and Euler's equation, the thrust $F$ and torque $M$ can be calculated from $u$.\\
\begin{equation}
\begin{split}
F = \frac{(ua_{z} + g) mass }{\cos \theta \cos \phi}
\end{split}
\end{equation}
\begin{equation}
\begin{split}
M = I \dot{\omega} + \omega \times I\omega
\end{split}
\end{equation}
The acceleration $ua$ and angular acceleration $\dot{\omega}$ of the quadcopter will be obtained by using the LQG controller, then the thrust $F$ and torque $M$ can be obtained via equations, because the estimated mass and inertia matrix need to be sent to the LQG controller in the online parameter estimation simulator and the estimated mass and inertia matrix will be used to get the thrust $F$ and torque $M$. The online parameter estimation simulator will be presented in the Section (\ref{Online parameter estimation simulation modelsection}).
\section{Parameter estimation}
\subsection{State space models for EM}
The state space model has been described in Section (\ref{ssm}) . It is assumed that the transition density and the measurement model depend on some parameters. So a change has been made: $p(x_{k}|x_{k-1})$ changes to $ p(x_{k}|x_{k-1}, \theta)$. $p(y_{k}|x_{k})$ changes to $p(y_{k}|x_{k},\theta)$. \\
The posterior distribution becomes:\\
\begin{equation}
p(x|y,\theta^{(i-1)}) = p(x_{1},...,x_{n} |y_{1},...,y_{n},\theta^{(i-1)})
\end{equation}
Extended Kalman-Rauch recursion based on the paper by Holmes\cite{term17} has been used to calculate the posterior distribution. This method is more efficient for state space models. Let $E^{(i-1)} = \int p(x|y,\theta^{(i-1)})$ and the joint log-likelihood based on the paper by Holmes\cite{term17} can be expressed as :\\

\begin{equation}
\begin{split}
\log p(x,y \mid \theta) &= \log p(x_{1}) \\
&\quad + \sum_{k=2}^{N} \log p(x_{k} \mid x_{k-1}, \theta) \\
&\quad + \sum_{k=1}^{N} \log p(y_{k} \mid x_{k}, \theta)
\end{split}
\end{equation}
Then the function for parameter update can be simplified to:\\
\begin{equation}
\begin{split}
L(\theta) &= \sum_{k=2}^{N} E^{(i-1)}\big[\log p(x_{k} \mid x_{k-1}, \theta)\big] \\
&\quad + \sum_{k=1}^{N} E^{(i-1)}\big[\log p(y_{k} \mid x_{k}, \theta)\big]
\end{split}
\end{equation}
\subsection{Expectation step}
In this step extended Kalman Rauch recursion has been applied to compute the $q$ distribution. In this part, the extended Kalman filter uses forward recursion to estimate all hidden states $X$. The results of the extended Kalman filter are then passed to the Rauch smoother. The Rauch smoother will work backwards to optimize the results.
\subsubsection{Extended Kalman filter}
In order to find the hidden $X$, it starts from t = 1 until N. $Y = \left\{y_{1},y_{2},y_{3},...,y_{n}\right\}$ is known. Extended Kalman filter has been used to filter out the noise from $Y$. The estimation $X = \left\{x_{1},x_{2},x_{3},...,x_{n}\right\}$. The output of the controller has been extracted from the simulation of a quadcopter. From the introduction of the extended Kalman filter above, it has:
\subsubsection{Prediction model}
\begin{equation}
\check{x}_{k} = \Phi_{k-1}\hat{x}_{k-1} + w_{k}, w_{k}\sim N(0,Q)
\end{equation}
$\Phi$ is the Jacobian matrix (\ref{jacobianF}) mentioned in Section 3.\\
The covariance of prior estimate $\check{P}_{k}$ at time $k$:\\
\begin{equation}
\check{P}_{k} = \hat{\Phi}_{k-1}\hat{P}_{k-1}\hat{\Phi}^{T}_{k-1} + Q_{k} 
\end{equation}
The observation equation is as follows:\\
\begin{equation}
y_{k} = H_{k}x_{k} + v_{k}, v_{k}\sim N(0,R)
\end{equation}
$H$ is known.
\subsubsection{Correction and update}
The extended Kalman filter gain matrix $K_{k}$ is:\\
\begin{equation}
K_{k} = \check{P}_{k} H^{T}_{k}(H_{k}\check{P}_{k}H^{T}_{k} + R_{k})^{-1}
\end{equation}
The covariance of the posterior estimate at time $k$ is:\\
\begin{equation}
\hat{P}_{k} = \check{P}_{k} - K_{k}H_{k} \check{P}_{k}
\end{equation}
Next, the extended Kalman gain is multiplied by the difference between the measured state $y_{k}$ and the predicted state $\check{x}_{k}$. The estimated state is given as:\\
\begin{equation}
\hat{x}_{k} = \check{x}_{k} + K_{k}[y_{k} - H_{k}\check{x}_{k}]
\end{equation}
\subsection{Rauch-Tung-Striebel smoother}
Next the Rauch-Tung-Striebel smoother works backwards from $k=N-1$ back to $k = 1$ to compute the estimation again. The RTS smoother algorithm is described in the book by Simo S{\"a}rkk{\"a}\cite{term12}. This smoother needs the estimation $X = \left\{\hat{x}_{1},\hat{x}_{2},\hat{x}_{3},...,\hat{x}_{n}\right\}$ and posterior covariance $P = \left\{\hat{P}_{1},\hat{P}_{2},\hat{P}_{3},...,\hat{P}_{n}\right\} $ from the extended Kalman filter. Initial value $\hat{x}_{s} = \hat{X}_{n},\hat{P}_s = \hat{P}_{n}$ .\\
\subsection{Prediction}
\begin{equation}
\check{x}_{p} = \hat{x}_{k} + \triangle t\dot{x}_{k}
\end{equation}
\begin{equation}
\check{P}_{p} =\hat{P}_{k} + Q 
\end{equation}
\subsection{Correction and update}
\begin{equation}
C_{k} = \frac{\hat{P}_{k}}{\check{P}_{p}}
\end{equation}
\begin{equation}
\hat{x}_{s} = \hat{x}_{k} + C_{k}(\hat{x}_{s} - \check{x}_{p}) 
\end{equation}
\begin{equation}
\hat{P}_{s} = \hat{P}_{k} + C_{k}(\hat{P}_{s} - \check{P}_{p})C_{k}^{T}   
\end{equation}
\subsection{Maximization step}
Now the objective function $L(\theta)$ will be maximized with respect to parameter $\theta$. \\
\begin{equation}
\begin{split}
L(\theta) &= \sum_{k=2}^{N} E^{(i-1)} \big[\log p(x_{k} \mid x_{k-1}, \theta)\big] \\
&\quad + \sum_{k=1}^{N} E^{(i-1)} \big[\log p(y_{k} \mid x_{k}, \theta)\big]
\end{split}
\end{equation}

The partial derivative of $L(\theta)$ with respect to parameter $\theta$ will be derived. Let the partial derivative equal zero. The corresponding parameters which let the objective function $L(\theta)$ get the maximum value will be obtained.\\
\begin{equation}
\frac{\partial L(\theta)}{\partial \theta} = 0
\end{equation}
\subsection{EM application-mass estimation}
The expectation maximization algorithm will be used to estimate the parameter mass based on observations. The acceleration of the z axis has been considered. $F$  has been defined as thrust of the quadcopter. $F_{k-1}$ depends on the state $x_{k-1}$. $r_{k-1}$ means the last element in the rotation matrix and it depends on state $x_{k-1}$. According to the state space model it results in:\\
\begin{equation}
\begin{aligned}
v_{k} = v_{k-1} + a\triangle t = v_{k-1} + (-g + \frac{F_{k-1}}{m}r_{k-1})\triangle t
\end{aligned}
\end{equation}
\begin{equation}
\begin{aligned}
y_{k} = v_{k} + \delta_{k} 
\end{aligned}
\end{equation}
$\delta_{k}$ is gaussian white noise. Our objective function $L$ results in:\\
\begin{equation}
\begin{split}
L(\theta) &= \sum_{k=2}^{n} E^{(i-1)} \big[\log p(v_{k} \mid v_{k-1}, \theta)\big] \\
&\quad + \sum_{k=1}^{n} E^{(i-1)} \big[\log p(y_{k} \mid v_{k}, \theta)\big]
\end{split}
\end{equation}

For SDE model, a random noise is added. The stochastic diﬀerential equation is:\\ 
\begin{equation}
\begin{aligned}
dv_{t} = (-g + \frac{F_{k-1}}{m}r_{k-1})dt + \delta dw_{t}
\end{aligned}
\end{equation}
$\delta$ is a scaling factor. For a short step $\triangle t$, the updated formula of the SDE model can be approximated as:\\
\begin{equation}
\begin{split}
v_{k} = v_{k-1} + (-g + \frac{F_{k-1}}{m}r_{k-1})\triangle t + \delta\sqrt{\triangle t }\epsilon_{k}  ,
\end{split}
\end{equation}
$\epsilon_{k} \sim N (0,1)$. The transition density of the model is:\\
\begin{equation}
\begin{split}
p(v_{k}|v_{k-1}) &= N((v_{k}|v_{k-1} + (-g + \frac{F_{k-1}}{m}r_{k-1})\triangle t , \delta^{2}\triangle t))\\
&= \frac{1}{\sqrt{2\pi \delta^{2}\triangle t}} e^{-\frac{1}{2\delta^{2}\triangle t }[v_{k} - v_{k-1} - (-g + \frac{F_{k-1}}{m}r_{k-1})\triangle t ]^{2}}
\end{split}
\end{equation}

If $\frac{\partial L(m)}{\partial m} = 0$, $m$ is derived as follows: \\
\begin{equation}
\begin{split}
m &= \frac{\sum_{k=2}^{n}(r_{k-1}F_{k-1})^{2}\triangle t}{\sum_{k=2}^{n}\big[ r_{k-1}F_{k-1} g \triangle t + (v_{k} - v_{k-1})r_{k-1}F_{k-1}\big]}
\end{split}\label{massestimation}
\end{equation}
After the estimate of mass has been obtained, it will be sent to E step to update the state estimation.
\subsection{EM application-inertia estimation}
In this section, the EM algorithm has been used to estimate the inertia matrix from the observations made. From Section (\ref{dynamic}), it is known that the inertia matrix is related to angular velocity. According to the state space model it results in:\\
\begin{equation}
\begin{aligned}
\omega_{k} = \omega_{k-1} + \dot{\omega}\triangle t 
\end{aligned}
\end{equation}
According to Euler's equation it results in:\\
\begin{equation}
\begin{split}
\dot{\omega} &= I^{-1}(M - \omega_{k-1}\times{I\omega_{k-1}})
\end{split}
\end{equation}
$I$ is the inertia matrx, and its expression is the equation (\ref{inertia})
\begin{equation}
\begin{split}
\left[
\begin{array}{ccc}
 \dot{\omega}_{x} \\
 \dot{\omega}_{y} \\
 \dot{\omega}_{z}
\end{array}
\right] =
\left[
\begin{array}{c}
\frac{1}{I_{xx}}M_{x} + \frac{I_{yy} - I_{zz}}{I_{xx}}\omega_{y}\omega_{z}\\
\frac{1}{I_{yy}}M_{y} + \frac{I_{zz} - I_{xx}}{I_{yy}}\omega_{x}\omega_{z}\\
\frac{1}{I_{zz}}M_{z} + \frac{I_{xx} - I_{yy}}{I_{zz}}\omega_{x}\omega_{y}
\end{array}
\right]
\end{split}
\end{equation}
Because it is assumed that the observation noise is independent of $\theta$, the objective function $L$ can be simplified as:\\
\begin{equation}
\begin{split}
L(\theta) = \sum_{k = 2}^{n}E^{(i-1)}\big [\log p(\omega_{k}|\omega_{k-1},\theta) \big ] 
\end{split}
\end{equation}
For a short period of time $\triangle t$, the updated formula of the SDE model can be approximated as:\\
\begin{equation}
\begin{split}
\omega_{k} = \omega_{k-1} + \dot{\omega} \triangle t + \delta\sqrt{\triangle t} \epsilon_{k} , \ \epsilon_{k} \sim N (0,1)
\end{split}
\end{equation} 
$\delta$ is scaling factor. So the transition density of the model can be written as:\\ 
\begin{equation}
\begin{split}
p(\omega_{k}|\omega_{k-1}) &= N(\omega_{k}|\omega_{k-1} + \dot{\omega}\triangle t , \delta^{2}\triangle t)\\
&= \frac{1}{\sqrt{2\pi \delta^{2}\triangle t}} e^{-\frac{1}{2\delta^{2}\triangle t }(\omega_{k} - \omega_{k-1} - \dot{\omega} \triangle t )^{2}}
\end{split}
\end{equation}

\begin{equation}
\begin{split}
E^{(i-1)}\big[\log p(\omega_{k} \mid \omega_{k-1}, \theta)\big] &= 
-\frac{1}{2} E^{(i-1)} \log(2 \pi \delta^{2} \triangle t) \\
&\quad - \frac{1}{2 \delta^{2} \triangle t} 
E^{(i-1)} \\
&\quad
\underbrace{\big[ (\omega_{k} - \omega_{k-1} - \dot{\omega} \triangle t)^2 \big]}_{\text{This part will be expanded}}
\end{split}
\end{equation}

$\omega_{x}$, $\omega_{y}$, $\omega_{z}$ will be analyzed separately. For $\omega_{x}$:\\
\begin{equation}
\begin{split}
(\omega_{x,k} - \omega_{x,k-1} - \dot{\omega}_{x} \triangle t)^{2} &= 
\omega_{x,k}^{2} - 2 \omega_{x,k} \omega_{x,k-1} + \omega_{x,k-1}^{2} \\
&\quad + (\dot{\omega}_{x} \triangle t)^{2} 
- 
\\
&\quad
2 (\omega_{x,k} - \omega_{x,k-1}) \dot{\omega}_{x} \triangle t
\end{split}
\end{equation}

When $\frac{\partial L(I_{xx})}{\partial I_{xx}}= 0$, we can get the expression for $I_{xx}$, which is displayed in detail in appendix.

\section{Simulation}
In this section, the simulation settings of the quadcopter and results of parameter estimation will be presented. This section has three subsections. Two simulation models are introduced in the first subsection. The second subsection shows the results of offline parameter estimation. Subsection 3 shows the online parameter estimation results. 
\subsection{Simulation models}
In this subsection, two types of simulation models will be presented. The first type is the offline parameter estimation simulation model. The second type is the online parameter estimation simulation model.
\subsubsection{Offline parameter estimation}
\begin{figure}[b]
	\centering
	\includegraphics[width=7.56cm,height=5.5cm]{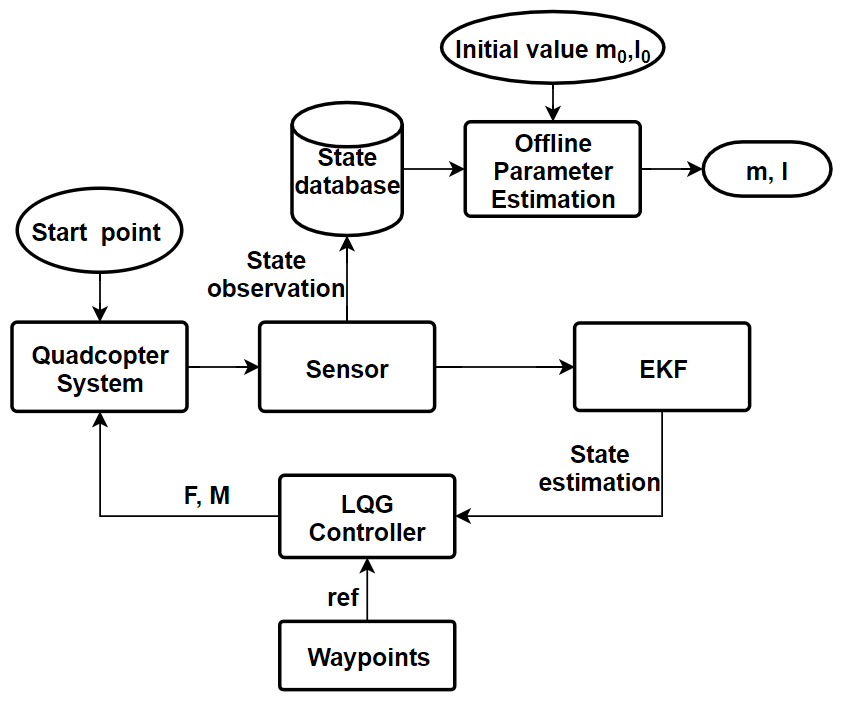}
	\caption{Offline parameter estimation simulation model} \label{Offline parameter estimation simulation model}
\end{figure}
The offline parameter estimation simulation model will be introduced in detail. As shown in figure (\ref{Offline parameter estimation simulation model}), in this model, all the parameters of the quadcopter are known. After the quadcopter flew a series of waypoints from the starting point, a series of state information have been obtained, which have been stored in the database. After observations have been obtained, the EM algorithm will be used to estimate the mass and inertia matrix based on the observations in the database.\\
A start point $[0.5, 1, 0]$ is given to the quadcopter system block. The sensor module outputs state observations to the state database. Then the EKF block recovers the state based on state observations. Regarding the sensor module, the parameter estimation results by using three different types of sensors in the following section will be presented. Waypoints are the desired position where the quadcopter should fly to. LQG controller outputs thrust $F$ and torque $M$ to the quadcopter system block. The quadcopter system will update the state. All the state observations will be stored in the state database block. Given the initial value mass $m_{0}$ and inertia matrix $I_{0}$, the offline parameter estimation block estimates mass and inertia matrix by EM algorithm iteratively and outputs the final estimated mass $m$ and inertia matrix $I$.
\subsubsection{Online parameter estimation}\label{Online parameter estimation simulation modelsection}
\begin{figure}[t]
	\centering
	\includegraphics[width=\columnwidth]{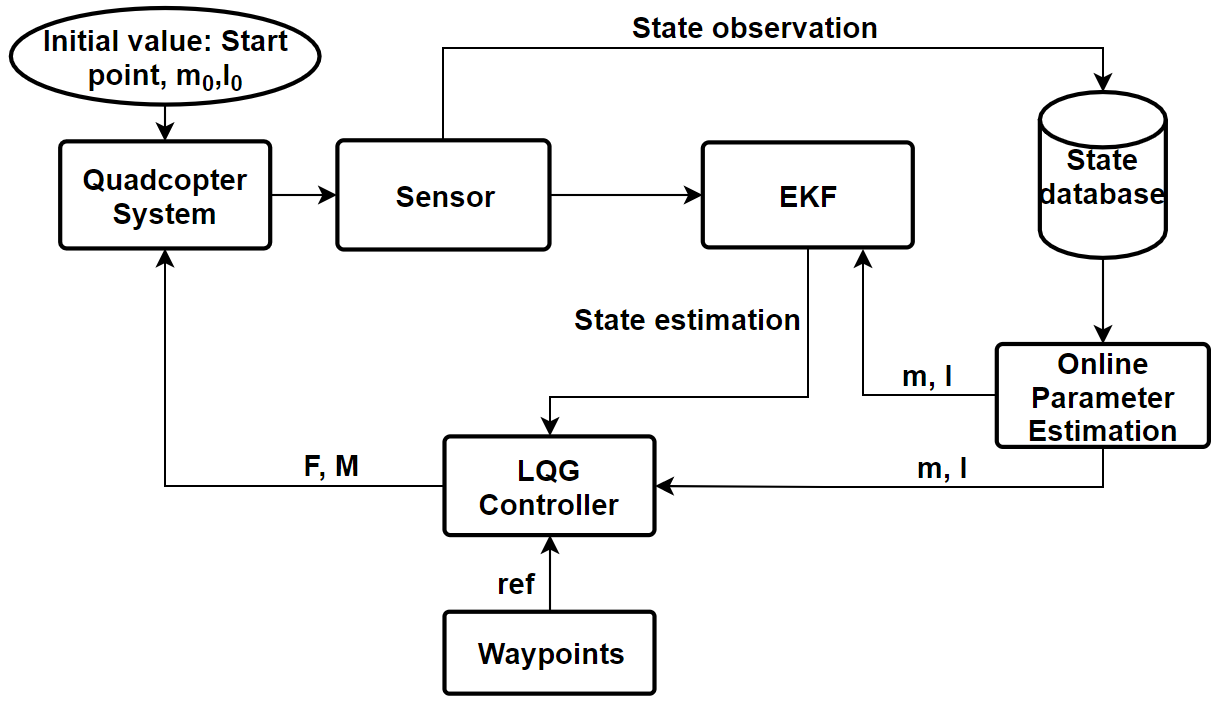}
	\caption{Online parameter estimation simulation model} \label{Online parameter estimation simulation model}
\end{figure}
In this part it is assumed that mass and inertia matrix are unknown, and the system block diagram has been given, as shown in figure (\ref{Online parameter estimation simulation model}). Figure (\ref{Online parameter estimation simulation model}) shows six modules, namely the quadcopter system module, the sensor module, the extended Kalman filter module, the data storage module and the online parameter estimation module.\\
Compared with the system in figure (\ref{Offline parameter estimation simulation model}), the difference of the system block diagram is that the online parameter estimation module has been embedded in the system. During the flight of the quadcopter, the system will automatically estimate the missing parameters and continue to optimize until convergence. As the length of the trajectory will increase with the increase of the number of waypoints, the calculation work of the online parameter estimation module will increase, so the online parameter estimation module is only allowed to select the latest 800 observations for parameter estimation. The working principle behind it is that before the program starts, the initial conditions have been given: the start point $[0.5, 1, 0]$, the initial values of mass $m_{0}$ and inertia matrix $I_{0}$. The online parameter estimation module extracts the latest 800 observations from the database for parameter estimation every 4 steps, and sends the estimated new mass $m$ and inertia matrix $I$ to the quadcopter system module. Then the entire system will run with the new parameters. For the sensor module, three different types of sensors will be tested and discussed too.\\
For the parameter estimation module, the initial value of mass $m_{0}$ and inertia matrix $I_{0}$ will be given as shown. The unit of the inertia matrix $I_{0}$ is $\text{kg} \cdot \text{m}^{2}$.\\
\begin{equation}
m_{0} = 0.001 \text{kg}
\end{equation}
\begin{equation}
I_{0}
=
\left[
\begin{array}{cccc}
1\text{E-}4 & 0 & 0\\
0 & 2 \text{E-}4 & 0\\
0 & 0 & 1 \text{E-}4
\end{array}
\right ] \label{initial_inertial}
\end{equation}
Covariance $P_{0}$, process noise covariance $Q_{0}$ and observed noise covariance $R_{0}$ are given as follows:\\
\begin{equation}
P_{0} = \text{diag}[1,1,1,1,1,1,1,1,1,1,1,1]
\end{equation}
\begin{equation}
Q_{0} = \text{diag}[1,1,1,1,1,1,1,1,1,1,1,1] \cdot 0.0707
\end{equation}
\begin{equation}
R_{0} = \text{diag}[1,1,1,1,1,1,1,1,1,1,1,1] \cdot 0.00707
\end{equation}

\subsection{Offline parameter estimation results}
The quadcopter flies over preset waypoints from the starting point $[0.5, 1, 0]$. The waypoints used in the simulation are $\left\{ [4, 3, 3], [3, 5, 4], [6, 4, 5], [4, 3, 4], [2, 1, 5] \right\}$. The corresponding parameter estimation results separately will be analyzed based on three different observation sources. There are three types of state observations for parameter estimation. The first type of state observation is obtained by EKF state estimation. In the second type, sensor B has been used to observe the full state of the quadcopter. Full state observation includes position, velocity, Euler angle and body angular velocity. In the third type of state observation, sensor C has been used to get the partial observation which only includes position, velocity and body angular velocity. Therefore, according to different observation sources, different estimation results have been obtained.\\
Due to the random noise in the quadcopter system and the observations, the parameter estimation results will be slightly different in different simulation cycles. Based on the same settings, 20 simulation cycles have been carried out and the resulting parameter estimates will be presented together.
\subsubsection{Based on state estimation from EKF }
\begin{figure}[t]
	\centering
	\includegraphics[width=10cm,height=5.5cm]{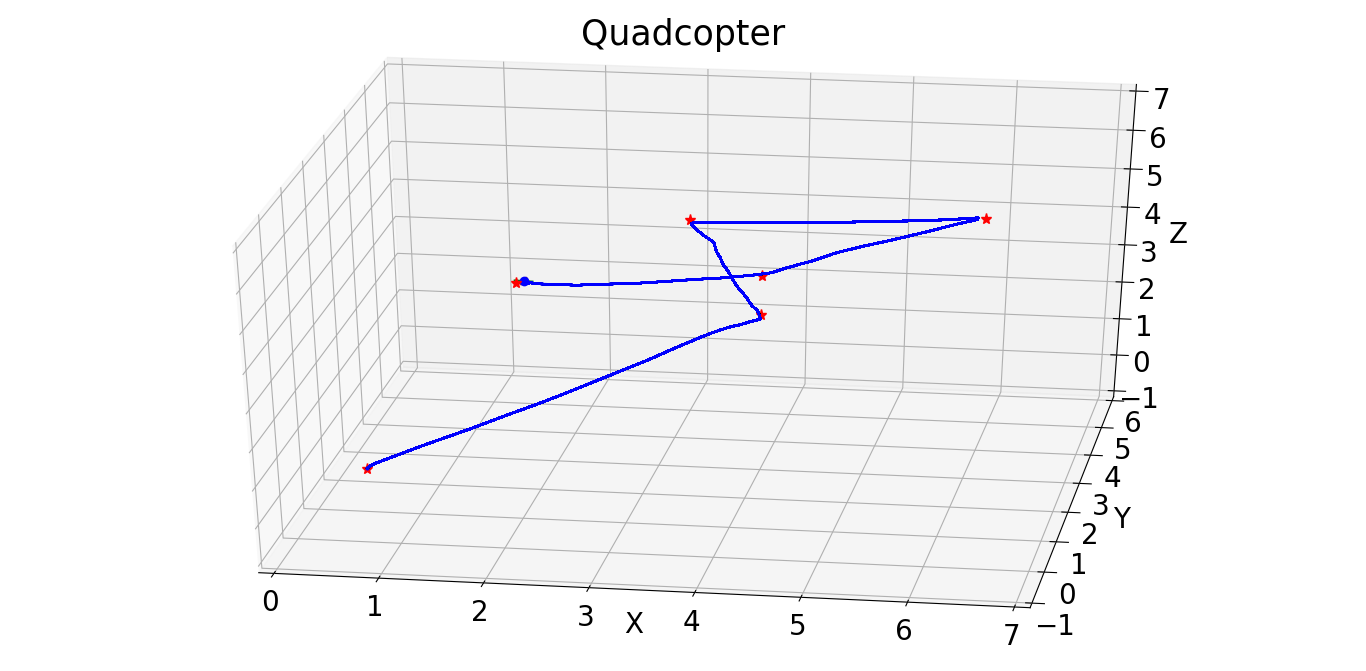}
	\caption{Quadcopter flight trajectory based on state estimation from EKF} \label{Quadcopter_EKF_off}
\end{figure} 
Figure (\ref{Quadcopter_EKF_off}) shows the trajectory of the simulated quadcopter flying under the control of the controller. It can be seen from the trajectory that the quadcopter successfully flew through the preset waypoints. The state estimates from the EKF have been extracted, then the quadcopter can use the EM algorithm to estimate the parameters based on all the states of the whole trajectory.
\begin{figure}[t]
	\centering
	\includegraphics[width=\columnwidth]{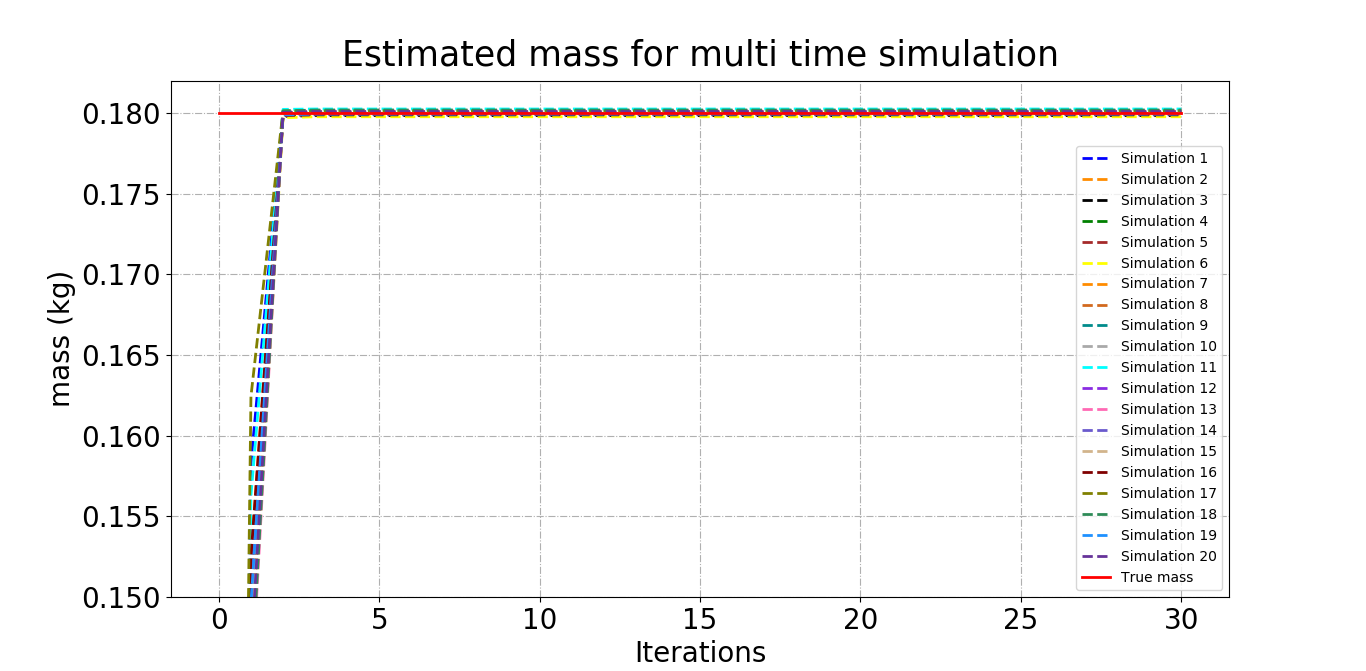}
	\caption{Estimated mass based on state estimation from EKF}\label{Estimated mass for multi time simulation_EKF_off}
\end{figure}
Next, the parameter estimation results will be presented. Figure (\ref{Estimated mass for multi time simulation_EKF_off}) shows the estimated results of the parameter mass for 20 cycles in dashed lines. The solid red line is the real parameter mass with a value of 0.18 kg. Overall, for different simulation cycles, the parameter mass converges to 0.18 kg rather quickly. By enlarging the curve, the parameter estimation results of 20 different simulation cycles fall within a certain range. This range is (0.1797,0.1802). The unit of mass is kg.
\begin{figure}[t]
	\centering
	\includegraphics[width=\columnwidth]{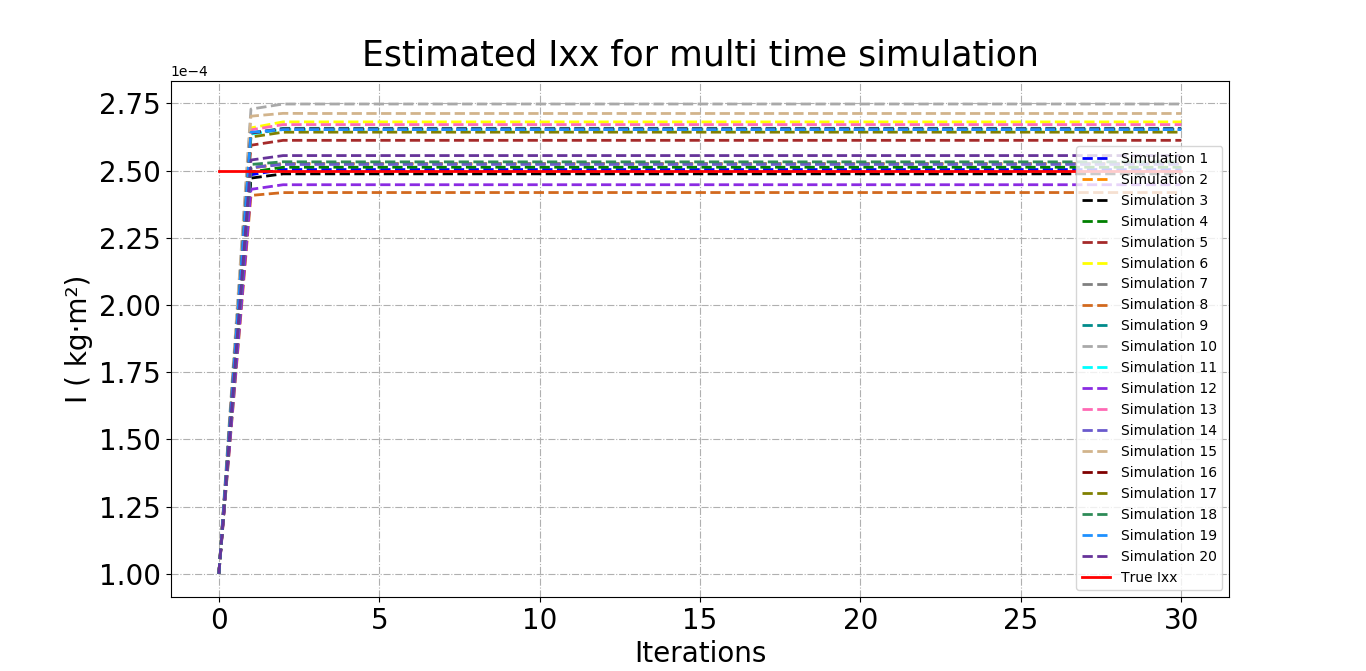}
	\caption{Estimated Ixx based on state estimation from EKF} \label{Estimated Ixx for multi time simulation_EKF_off}
\end{figure} 
Figure (\ref{Estimated Ixx for multi time simulation_EKF_off}) is the estimated result of the parameter $I_{xx}$. All dashed lines are the estimated results of $I_{xx}$, and the red solid line is the true value $I_{xx}$ 2.5E-4 $\text{kg} \cdot \text{m}^{2}$. It can be seen that the convergence value fluctuates within the range (2.418E-4, 2.75E-4) $\text{kg} \cdot \text{m}^{2}$.
\begin{figure}[t]
	\centering
	\includegraphics[width=\columnwidth]{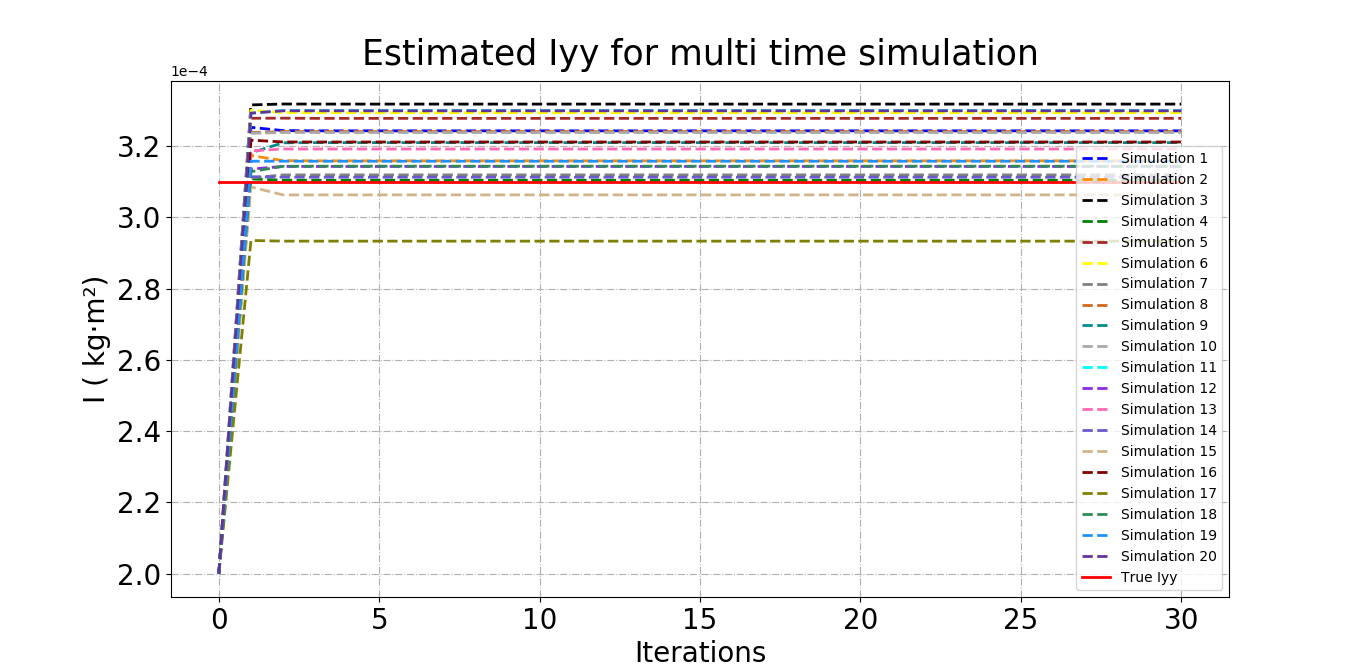}
	\caption{Estimated Iyy based on state estimation from EKF} \label{Estimated Iyy for multi time simulation_EKF_off}
\end{figure} 
Figure (\ref{Estimated Iyy for multi time simulation_EKF_off}) is the estimated result of the parameter $I_{yy}$. The red solid line is the true $I_{yy}$ 3.1E-4 $\text{kg} \cdot \text{m}^{2}$. All dashed lines are estimated values of $I_{yy}$. It can be seen that the convergence value fluctuates within the range (2.93E-4, 3.32E-4) $\text{kg} \cdot \text{m}^{2}$.
\begin{figure}[t]
	\centering
	\includegraphics[width=\columnwidth]{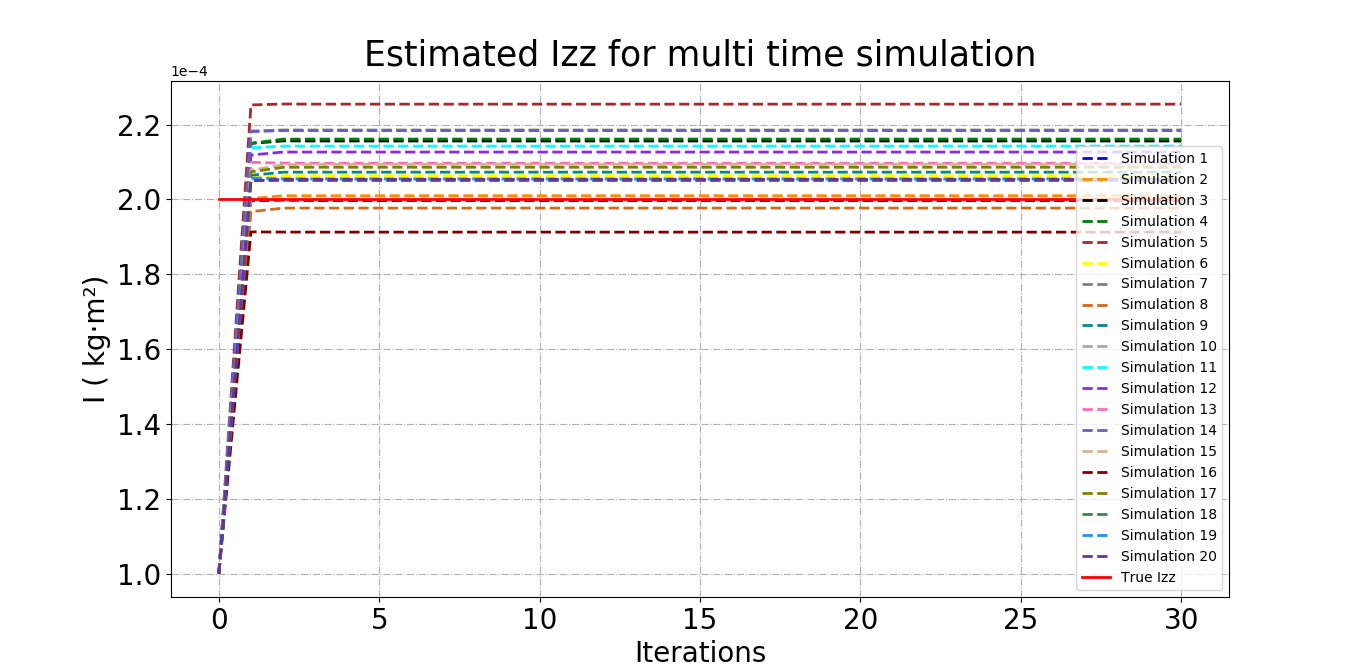}
	\caption{Estimated Izz based on state estimation from EKF} \label{Estimated Izz for multi time simulation_EKF_off}
\end{figure} 
The estimated $I_{zz}$ is shown in figure (\ref{Estimated Izz for multi time simulation_EKF_off}). The red solid line represents the real value of $I_{zz}$ 2E-4 $\text{kg} \cdot \text{m}^{2}$. All dashed lines are estimated values of $I_{zz}$. The estimated $I_{zz}$ based on 20 simulation cycles constitutes the range (1.912E-4, 2.255E-4) $\text{kg} \cdot \text{m}^{2}$ of $I_{zz}$. 
\subsubsection{Based on full state observation}
\begin{figure}[t]
	\centering
	\includegraphics[width=\columnwidth]{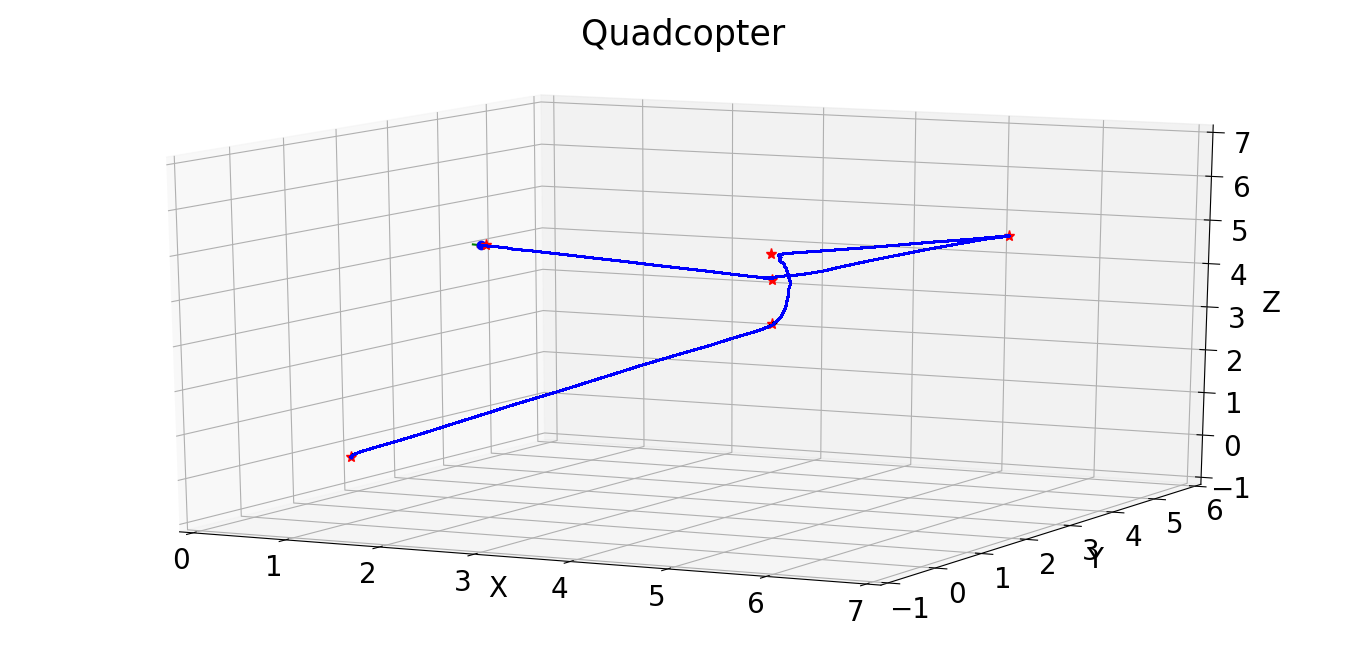}
	\caption{Quadcopter flight trajectory  based on full state observation} \label{Quadcopter_full_off}
\end{figure}

Figure (\ref{Quadcopter_full_off}) is the flight trajectory based on full state observation. Since the parameter estimation module is offline, all parameters of the quadcopter are known during flight. Compared with the trajectory in figure (\ref{Quadcopter_EKF_off}), the trajectory is not much different from the previous trajectory.
\begin{figure}[t]
	\centering
	\includegraphics[width=\columnwidth]{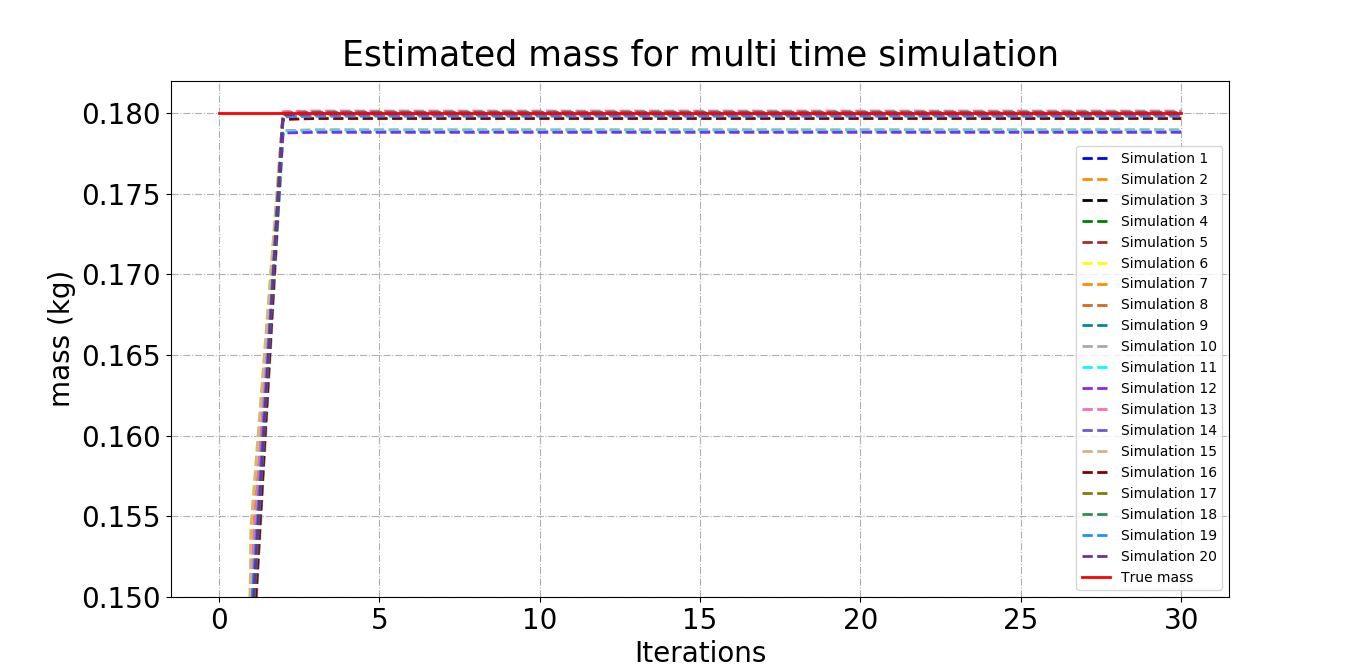}
	\caption{Estimated mass based on full state observation}\label{Estimated mass for multi time simulation_full_off}
\end{figure}
Figure (\ref{Estimated mass for multi time simulation_full_off}) shows the estimated parameter mass. In this figure, the red solid line is the real value of mass and the real mass is 0.18 kg. All the dashed lines are the estimated mass based on 20 simulation cycles. The estimated mass converged very fast in around 4th iteration. The range of the estimated mass is (0.1788, 0.1801) $\text{kg}$.
\begin{figure}[t]
	\centering
	\includegraphics[width=\columnwidth]{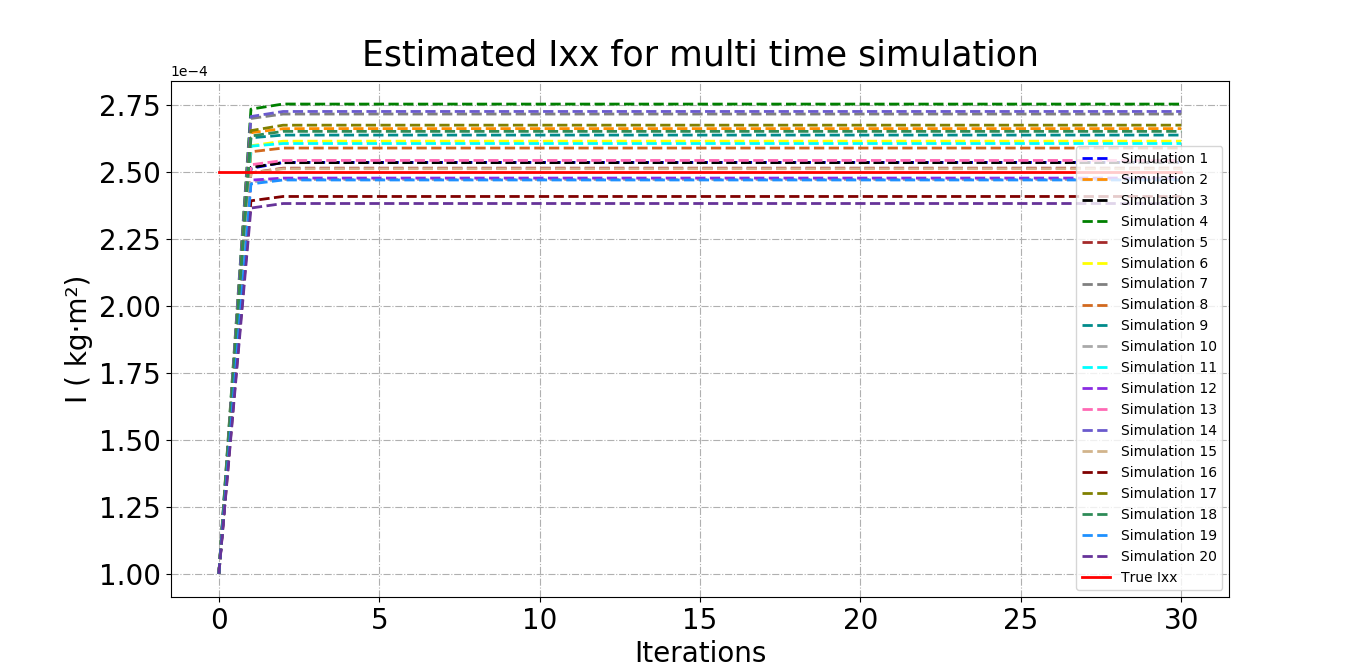}
	\caption{Estimated Ixx based on full state observation} \label{Estimated Ixx for multi time simulation_full_off}
\end{figure} 
The estimated parameter $I_{xx}$ is shown in figure (\ref{Estimated Ixx for multi time simulation_full_off}). The true value of $I_{xx}$ is 2.5E-4 $\text{kg} \cdot \text{m}^{2}$, which is labeled by the red solid line. All the dashed lines are the resuts of estimated $I_{xx}$. It can be seen from figure (\ref{Estimated Ixx for multi time simulation_full_off}) that the estimated $I_{xx}$ converges around the true value within a range of (2.383E-4, 2.755E-4) $\text{kg} \cdot \text{m}^{2}$.
\begin{figure}[b]
	\centering
	\includegraphics[width=\columnwidth]{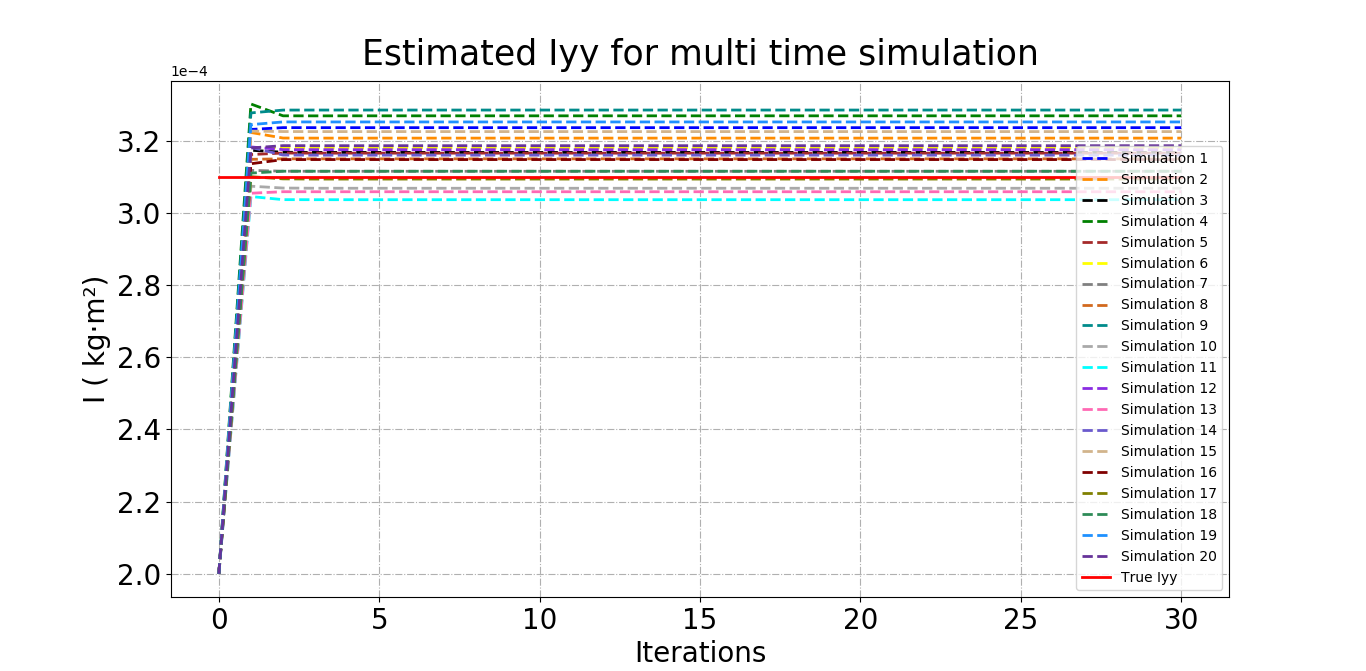}
	\caption{Estimated Iyy based on full state observation} \label{Estimated Iyy for multi time simulation_full_off}
\end{figure} 
Figure (\ref{Estimated Iyy for multi time simulation_full_off}) is the results of estimated $I_{yy}$. The red solid line is the real value of $I_{yy}$ 3.1E-4 $\text{kg} \cdot \text{m}^{2}$. All the dashed lines converge around the red solid line. Those dashed lines represent the estimated $I_{yy}$. The range of estimated $I_{yy}$ is (3.037E-4, 3.286E-4). The unit of $I_{yy}$ is $\text{kg} \cdot \text{m}^{2}$.
\begin{figure}[b]
	\centering
	\includegraphics[width=\columnwidth]{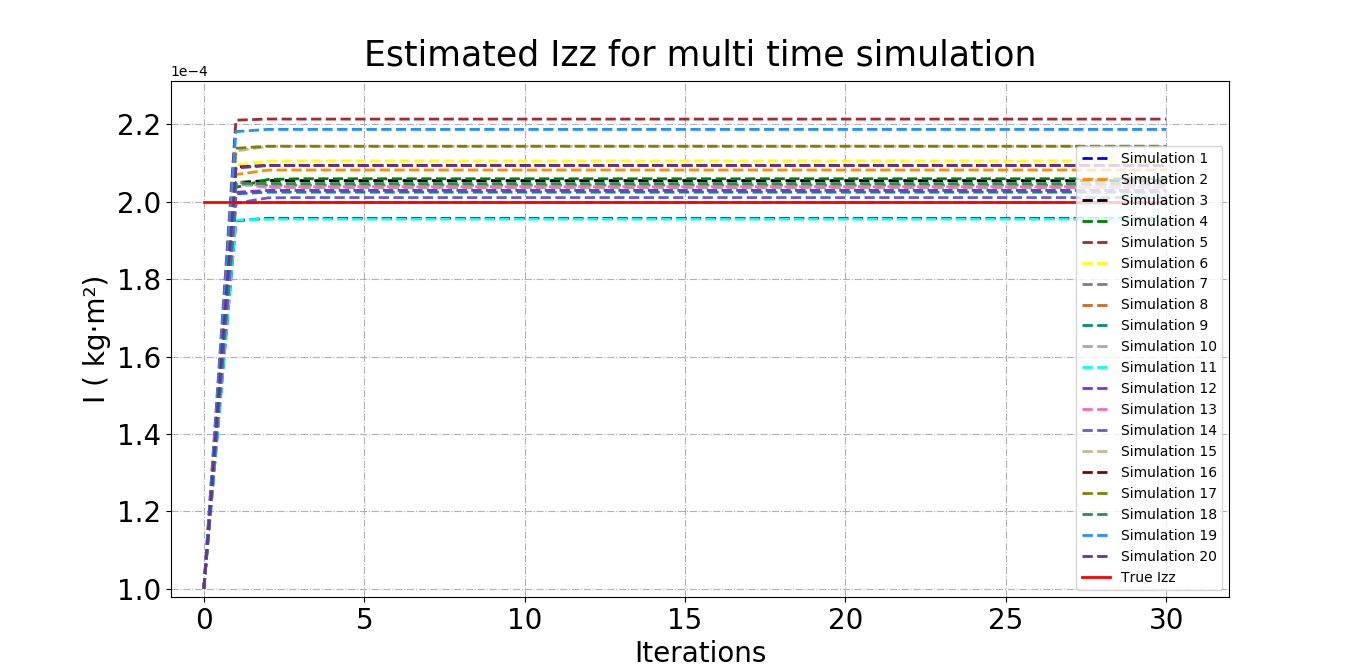}
	\caption{Estimated Izz based on full state observation} \label{Estimated Izz for multi time simulation_full_off}
\end{figure} 
Figure (\ref{Estimated Izz for multi time simulation_full_off}) shows the results of estimated $I_{zz}$. The red solid line represents the true value $I_{zz}$ 2E-4 $\text{kg} \cdot \text{m}^{2}$. All the dashed lines are the estimated $I_{zz}$ for 20 simulation cycles. The estimated results are in the range (1.955E-4, 2.214E-4). The unit of $I_{zz}$ is $\text{kg} \cdot \text{m}^{2}$.
\subsubsection{Based on partial state observation}
\begin{figure}[t]
	\centering
	\includegraphics[width=\columnwidth]{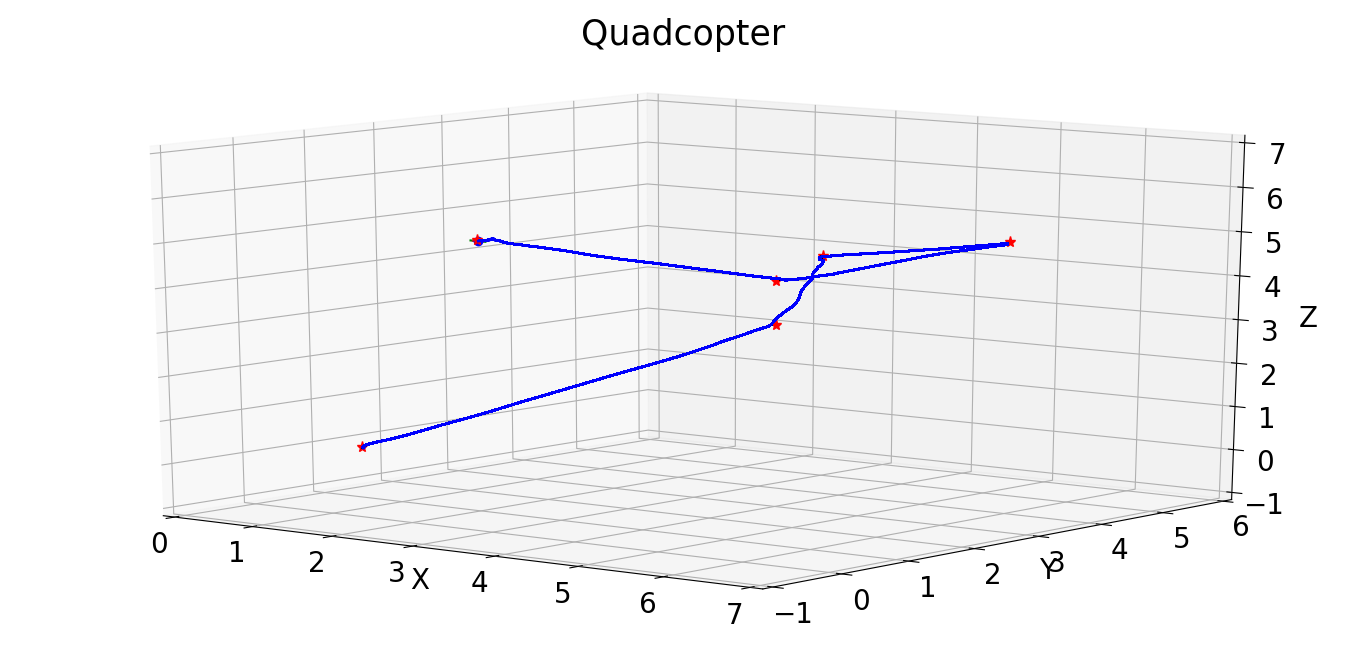}
	\caption{Quadcopter flight trajectory  based on partial state observation} \label{Quadcopter_part_off}
\end{figure} 
The trajectory based on partial state observation is shown in figure (\ref{Quadcopter_part_off}). The partial observations will be used to estimate the parameters mass and inertia matrix.
\begin{figure}[t]
	\centering
	\includegraphics[width=\columnwidth]{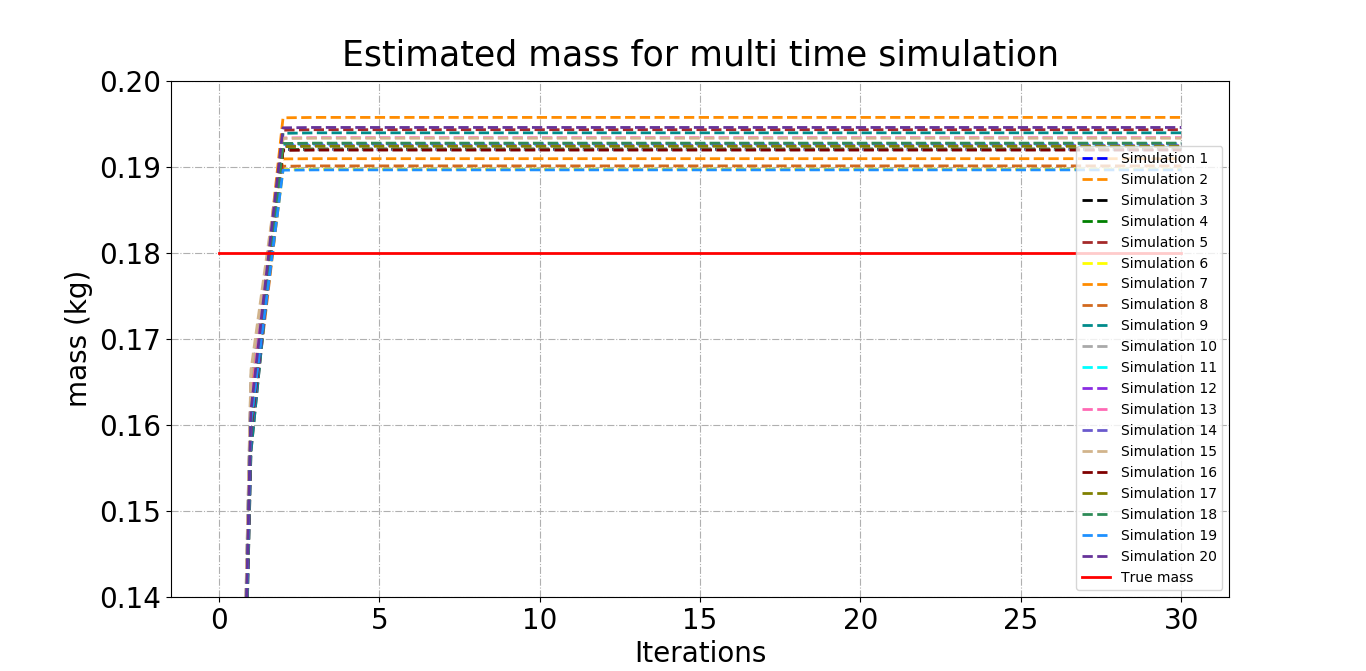}
	\caption{Estimated mass based on partial state observation}\label{Estimated mass for multi time simulation_part_off}
\end{figure}
The results of estimated mass is shown in figure (\ref{Estimated mass for multi time simulation_part_off}). The red solid line is the true value of mass. All the dashed lines are the results of estimated mass. It can be seen that the converged value is still close to the true value 0.18 kg. For 20 simulation cycles, the upper bound is 0.196 kg and the lower bound is 0.1896 kg. All the estimated mass results from 20 simulation cycles converge in the range (0.1896, 0.196) kg.
\begin{figure}[b]
	\centering
	\includegraphics[width=\columnwidth]{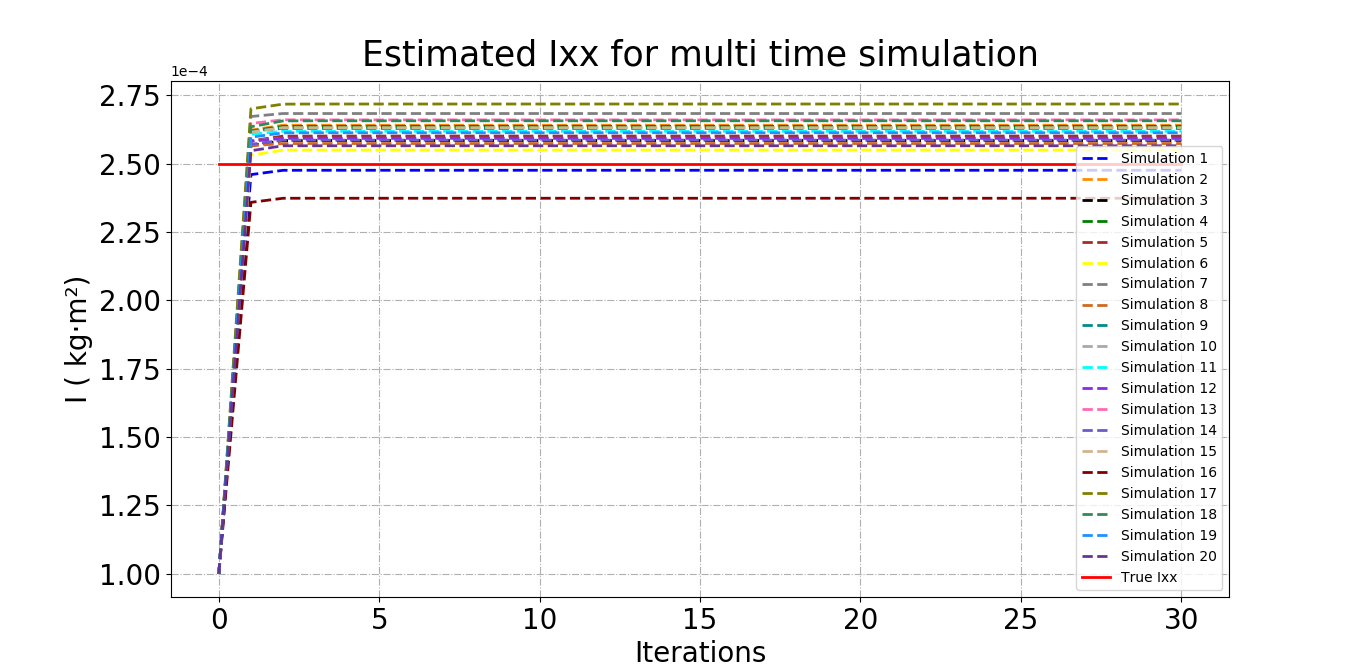}
	\caption{Estimated Ixx based on partial state observation} \label{Estimated Ixx for multi time simulation_part_off}
\end{figure} 
The estimated  $I_{xx}$ based on 20 simulation cycles is shown in figure (\ref{Estimated Ixx for multi time simulation_part_off}). The true value of $I_{xx}$ is 2.5E-4 $\text{kg} \cdot \text{m}^{2}$ which is labeled by the red solid line in this figure. All the dashed lines are distributed around the red solid line. The range is (2.373E-4, 2.718E-4) $\text{kg} \cdot \text{m}^{2}$.
\begin{figure}[b]
	\centering
	\includegraphics[width=\columnwidth]{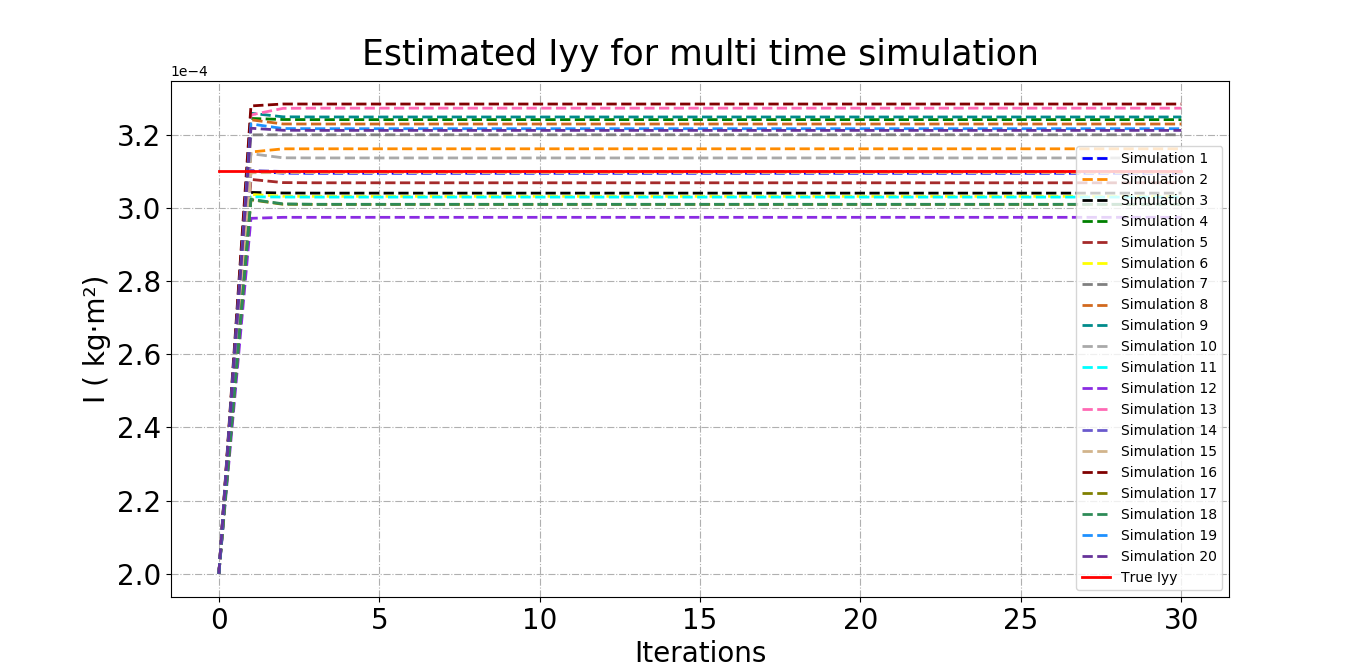}
	\caption{Estimated Iyy based on partial state observation} \label{Estimated Iyy for multi time simulation_part_off}
\end{figure} 
Figure (\ref{Estimated Iyy for multi time simulation_part_off}) shows the results of estimated $I_{yy}$. The red solid line is the true $I_{yy}$ and the true $I_{yy}$ is 3.1E-4 $\text{kg} \cdot \text{m}^{2}$. It can be seen that all the dashed lines are distributed around the red solid line. In 20 simulation cycles, the convergence range of $I_{yy}$ is (2.974E-4, 3.284E-4) $\text{kg} \cdot \text{m}^{2}$.
\begin{figure}[t]
	\centering
	\includegraphics[width=\columnwidth]{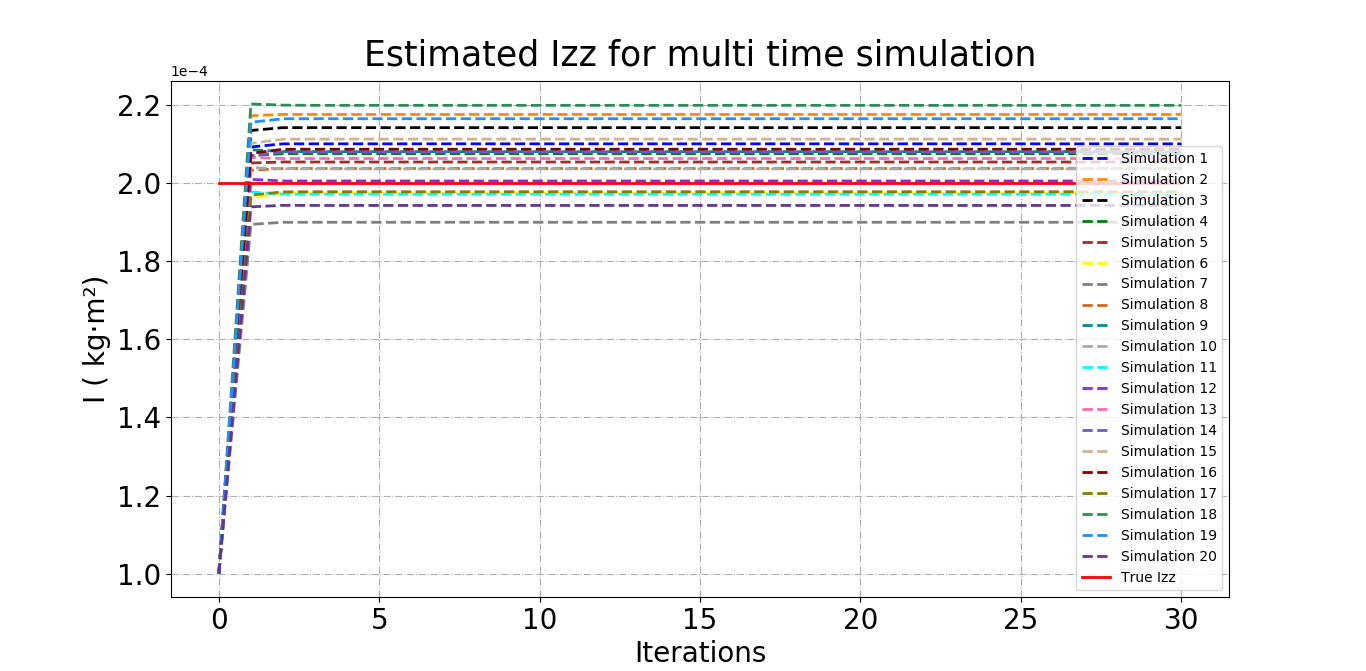}
	\caption{Estimated Izz based on partial state observation} \label{Estimated Izz for multi time simulation_part_off}
\end{figure} 
Figure (\ref{Estimated Izz for multi time simulation_part_off}) shows the results of estimated $I_{zz}$ based on 20 simulation cycles. The red solid line represents real $I_{zz}$ and the value of real $I_{zz}$ is 2E-4 $\text{kg} \cdot \text{m}^{2}$. All the dashed lines are  distributed around the red line. In 20 simulation cycles, the convergence range of $I_{zz}$ is (1.898E-4, 2.198E-4) $\text{kg} \cdot \text{m}^{2}$.
\subsubsection{Summary}
The obtained estimation results have been organized into tables (\ref{Offline estimated mass}) and (\ref{Offline estimated inertia matrix}). The estimated parameter results based on observations from different sources will be compared. \\
Table (\ref{Offline estimated mass}) lists the estimated parameter mass based on different types of observations. The second row is the true value of mass. The third row to the fifth row represent the estimated mass based on three different observations. This table includes the value range calculated out of 20 simulation cycles. Compared with the real value, it is found that the parameter mass estimation based on the EKF state estimation and the full state observation are closer to the real value. The estimated mass range based on EKF state estimation is 0.17\% lower than the real mass and 0.11\% higher than the real mass. The estimated mass range based on full state observation is 0.67\% lower than the real mass and 0.06\% higher than the real mass. However, the estimation of the parameter mass based on partial observations is somewhat inaccurate. The estimated mass range based on partial observation is 5.33\% to 8.89\% higher than the real mass. This is due to the fact that the Euler angle is not included in partial observations, which can cause a deviation. From formula (\ref{massestimation}), it can be seen that mass estimation needs the Euler angle.\\
\begin{table}[h]
	\renewcommand\arraystretch{1.5}
	\centering
	\setlength{\abovecaptionskip}{0pt}%
	\setlength{\belowcaptionskip}{10pt}%
	\caption{Offline estimated mass}
	\begin{tabular}{c|c}
		\hline  
		& Mass (kg) \\ 
		\hline  
		Real value	 & 0.18  \\
		\hline 
		State estimation from EKF & (0.1797, 0.1802)   \\
		\hline 
		Full state observation	& (0.1788, 0.1801)   \\
		\hline 
		Partial state observation & (0.1896, 0.196)  \\
		\hline 
	\end{tabular}\label{Offline estimated mass}
\end{table}
\begin{table*}[h]
	\renewcommand\arraystretch{1.5}
	\centering
	\setlength{\abovecaptionskip}{0pt}%
	\setlength{\belowcaptionskip}{10pt}%
	\caption{Offline estimated inertia matrix }
	\begin{tabular}{p{3cm}|c|c|c}
		\hline  
		& $I_{xx} (\text{kg} \cdot \text{m}^{2})$ & $I_{yy} (\text{kg} \cdot \text{m}^{2})$  & $I_{zz} (\text{kg} \cdot \text{m}^{2})$ \\ 
		\hline  
		Real value	  & 2.5E-4  & 3.1E-4	&2E-4\\
		\hline 
		EKF state estimation  & (2.418E-4, 2.75E-4) & (2.93E-4, 3.32E-4) & (1.912E-4, 2.255E-4)  \\
		\hline 
		Full state observation	 & (2.383E-4, 2.755E-4)& (3.037E-4, 3.286E-4) & (1.955E-4, 2.214E-4)  \\
		\hline 
		Partial state observation & (2.373E-4, 2.718E-4)  & (2.974E-4, 3.284E-4) & (1.898E-4, 2.198E-4) \\
		\hline 
	\end{tabular}\label{Offline estimated inertia matrix}
\end{table*}
Table (\ref{Offline estimated inertia matrix}) shows the estimated parameter inertia matrix based on 20 simulation cycles. The second row of the table is the true value of the inertia matrix. The third to fifth rows are the estimated inertia matrix results based on three different types of observations. This table lists the value range based on 20 simulation cycles in the third to fifth rows rather than a single value. The estimated $I_{xx}$ range based on EKF state estimation is 3.28\% lower than the real $I_{xx}$ and 10\% higher than the real $I_{xx}$. The estimates of the three components of the inertia matrix are quite close the three components of the actual inertia matrix. This is because the estimation of the inertia matrix is related to the angular velocity, and the angular velocity can be observed in three different types of observations.

\subsection{Online parameter estimation results}
In this section, the online parameter estimation results based on three different observations will be presented. Unlike the offline parameter estimation, the online parameter estimation module is embedded in the quadcopter. The preset waypoints are the same as before. In the online parameter estimation module, the EM algorithm will be used to perform parameter estimation every four steps based on existing observations rather than the whole trajectory in the end. As the waypoints increase and the trajectory becomes longer, the calculation of parameter estimates will increase and affect flight. Therefore, the number of observations used to estimate the parameters should be limited. In this system, the online parameter estimation module is only allowed to use up to 800 observations to estimate parameter mass and inertia matrix. As the quadcopter flies and the observations increase, the past observations will be deleted and only the most recent 800 observations will be used by the online parameter estimation module. Under the same settings, the quadcopter simulator based on online parameter estimation runs 20 times. Based on these 20 simulation cycles, 20 sets of parameter estimation results have been obtained. Next, the simulation results based on three different types of observations will be presented.
\subsubsection{Based on state estimation from EKF }
\begin{figure}[t]
	\centering
	\includegraphics[width=\columnwidth]{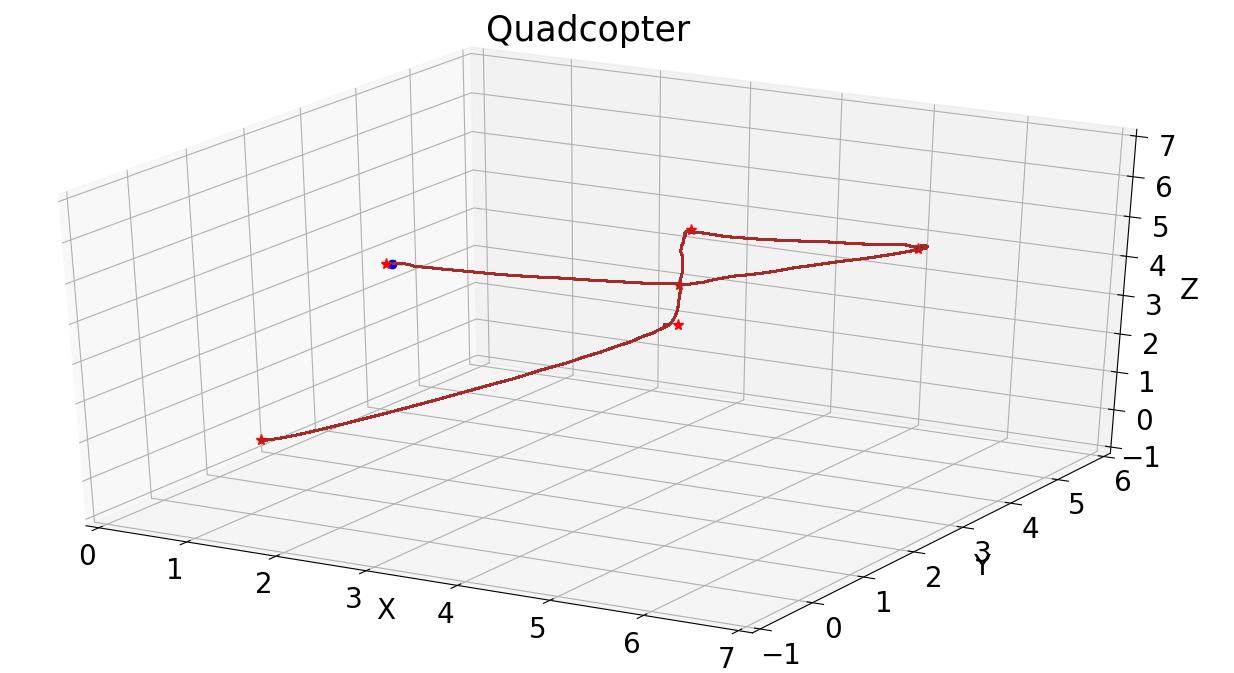}
	\caption{Quadcopter flight trajectory based on state est from EKF} \label{Quadcopter_EKF_on}
\end{figure}
Figure (\ref{Quadcopter_EKF_on}) shows the trajectory based on state estimation from EKF. Compared to the trajectory in figure (\ref{Quadcopter_EKF_off}), the trajectory between the waypoint $[0.5, 1, 0]$ and the waypoint $[4, 3, 3]$ in figure (\ref{Quadcopter_EKF_on}) is a little concave downward.
\begin{figure}[t]
	\centering
	\includegraphics[width=\columnwidth]{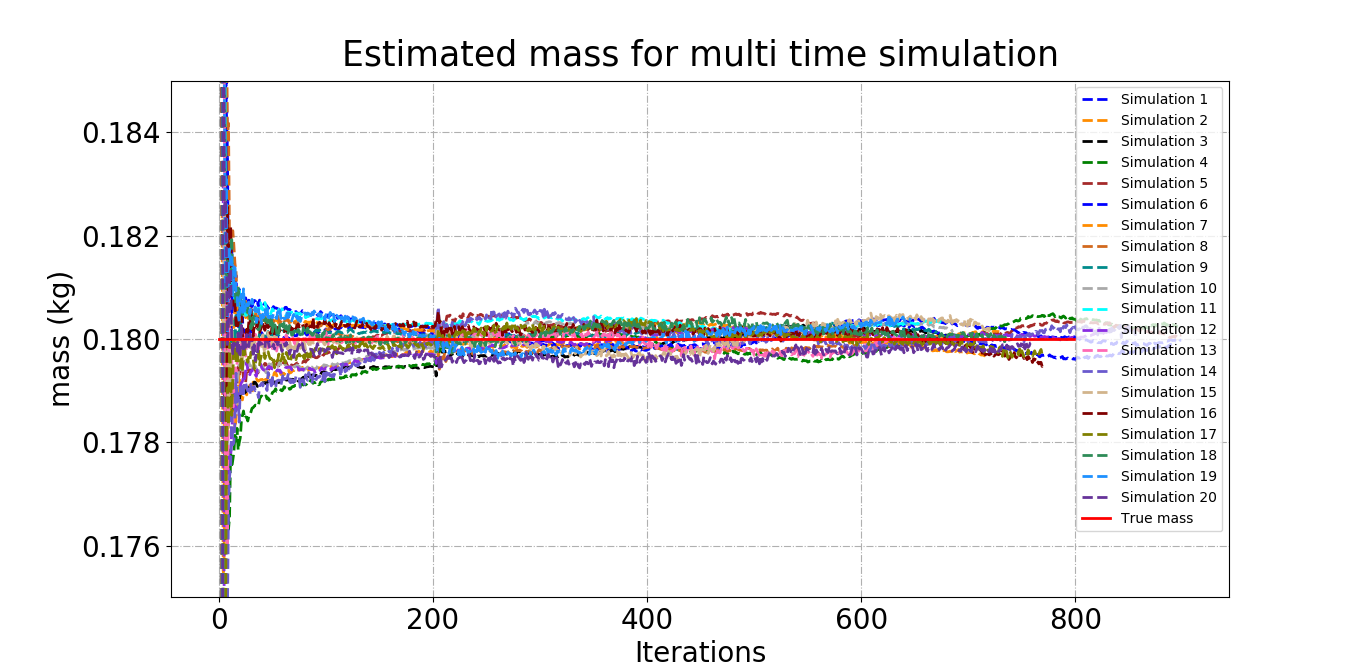}
	\caption{Estimated mass based on state est from EKF}\label{Estimated mass for multi time simulation_EKF_on}
\end{figure}
Figure (\ref{Estimated mass for multi time simulation_EKF_on}) shows the results of the estimated mass. The red solid line shows the real mass 0.18 kg. All the dashed lines represent the results of the estimated mass. It can be seen that the estimated mass from different simulation cycles distributes around the red solid line. The range of the convergence values is (0.1795, 0.1806) kg.
\begin{figure}[t]
	\centering
	\includegraphics[width=\columnwidth]{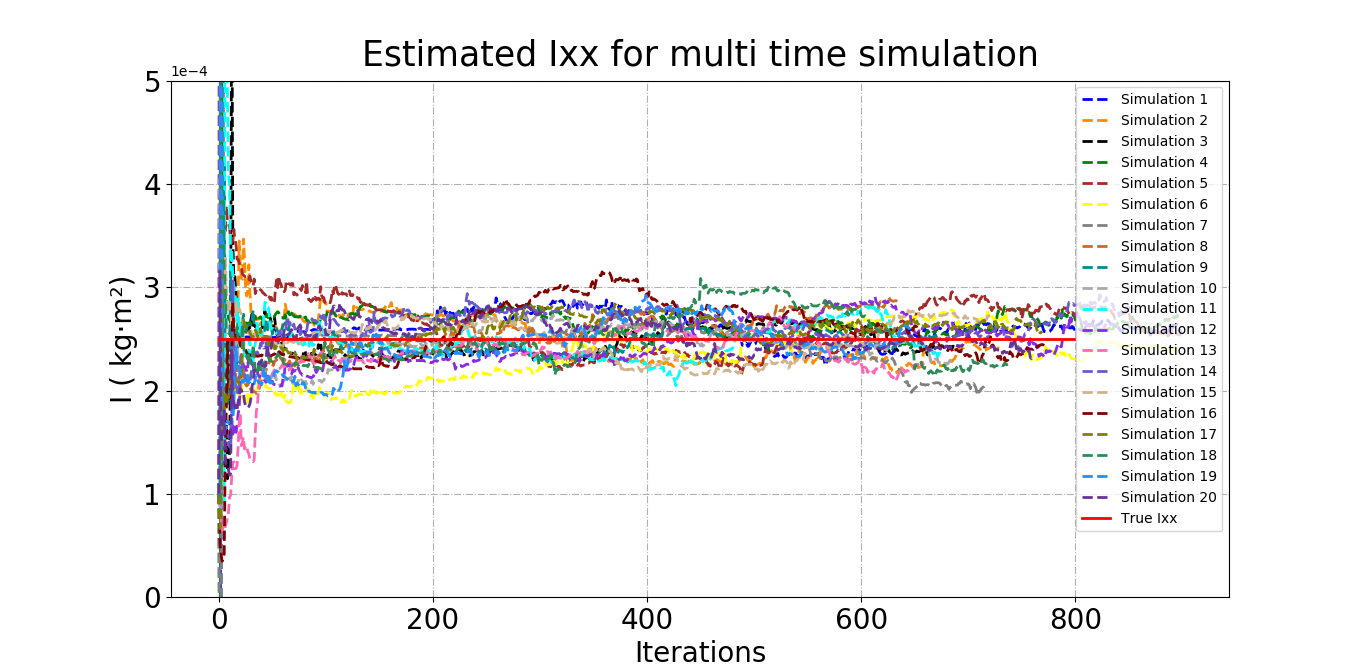}
	\caption{Estimated Ixx based on state est from EKF} \label{Estimated Ixx for multi time simulation_EKF_on}
\end{figure} 
Figure (\ref{Estimated Ixx for multi time simulation_EKF_on}) shows the results of estimated $I_{xx}$. The true value of $I_{xx}$ is 2.5E-04 $\text{kg} \cdot \text{m}^{2}$ which is represented by red solid line. All the dashed lines are the results of estimated $I_{xx}$. In general, the results of all simulation cycles gradually converge to the true value. The convergence range is (2E-4, 3.15E-4) $\text{kg} \cdot \text{m}^{2}$.
\begin{figure}[t]
	\centering
	\includegraphics[width=\columnwidth]{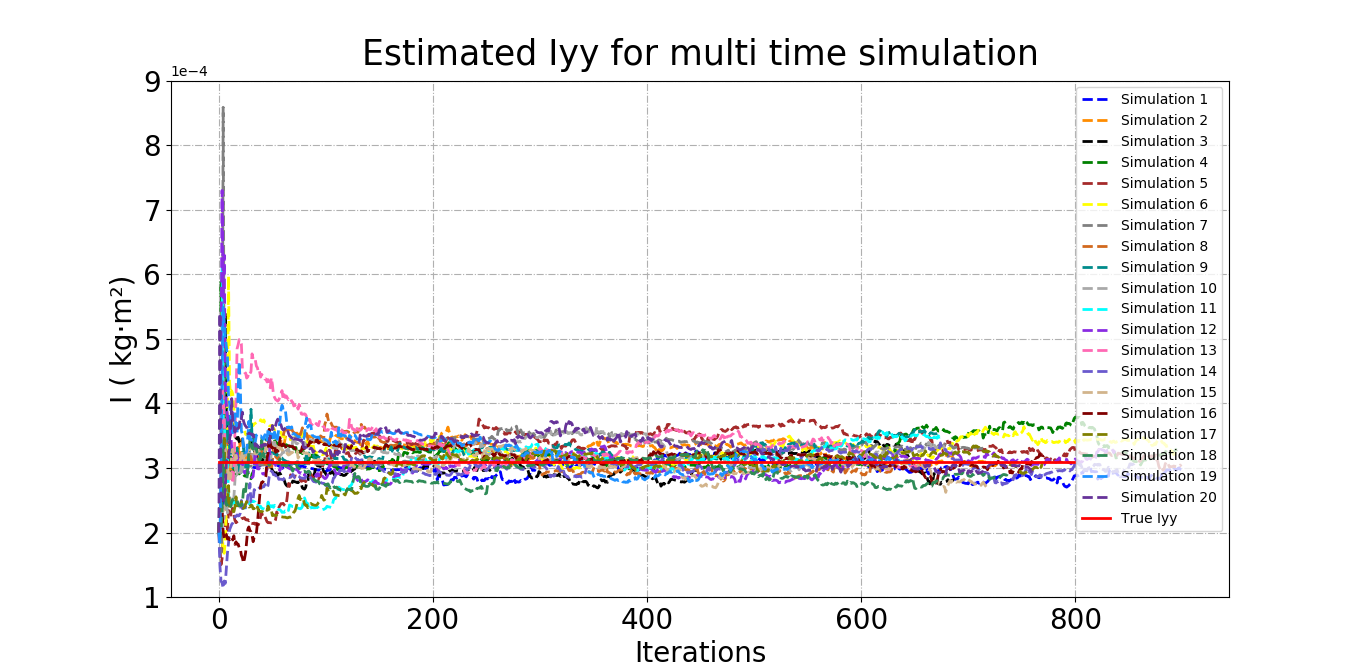}
	\caption{Estimated Iyy based on state est from EKF} \label{Estimated Iyy for multi time simulation_EKF_on}
\end{figure} 
Figure (\ref{Estimated Iyy for multi time simulation_EKF_on}) shows results of estimated $I_{yy}$. The red solid line is true value of $I_{yy}$ 3.1E-4 $\text{kg} \cdot \text{m}^{2}$. All the dashed lines fluctuate around the red solid line. It can be seen that the estimation of the parameter $I_{yy}$ gradually converges around the 200th iteration. The range of convergence value is (2.65E-4, 3.75E-4) $\text{kg} \cdot \text{m}^{2}$.
\begin{figure}[t]
	\centering
	\includegraphics[width=\columnwidth]{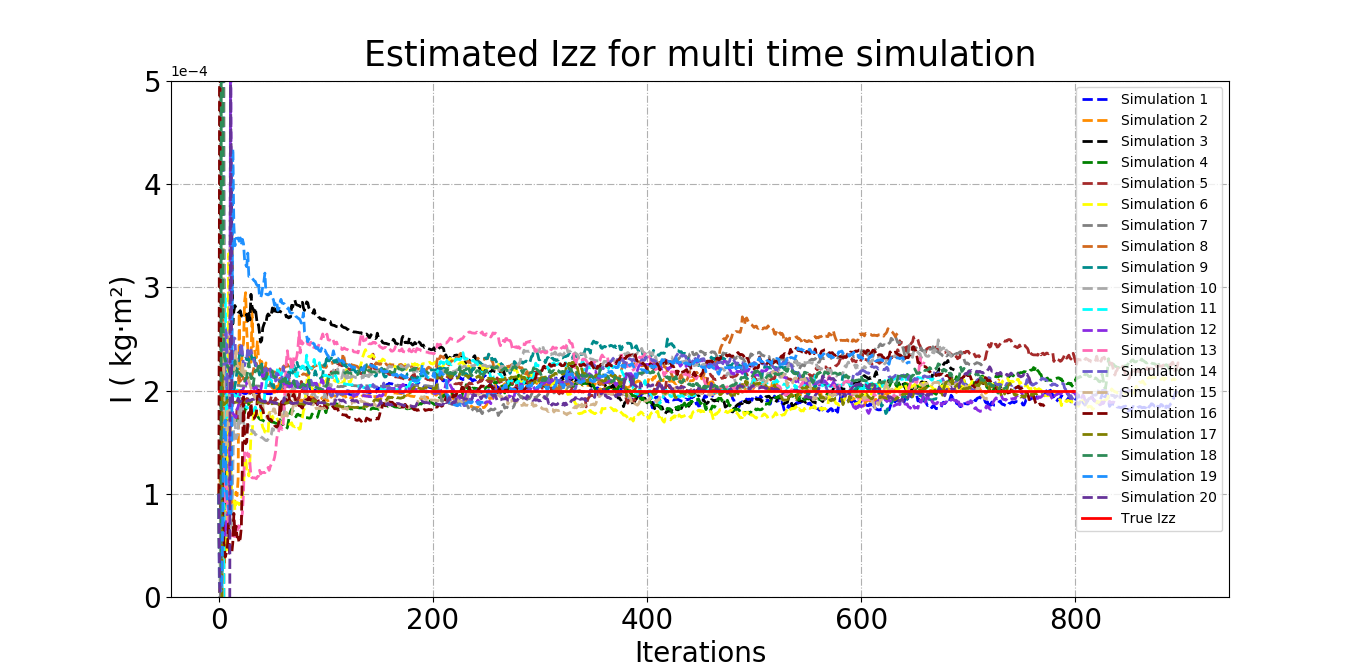}
	\caption{Estimated Izz based on state est from EKF} \label{Estimated Izz for multi time simulation_EKF_on}
\end{figure} 
The estimated $I_{zz}$ is shown in figure (\ref{Estimated Izz for multi time simulation_EKF_on}). The true value of $I_{zz}$ 2E-4 $\text{kg} \cdot \text{m}^{2}$ is marked by the solid red line. All dashed lines are the estimated $I_{zz}$. All dashed lines are all distributed near the red solid line. The range of convergence value is (1.7E-4, 2.6E-4) $\text{kg} \cdot \text{m}^{2}$.
\subsubsection{Based on full state observation}
\begin{figure}[t]
	\centering
	\includegraphics[width=\columnwidth]{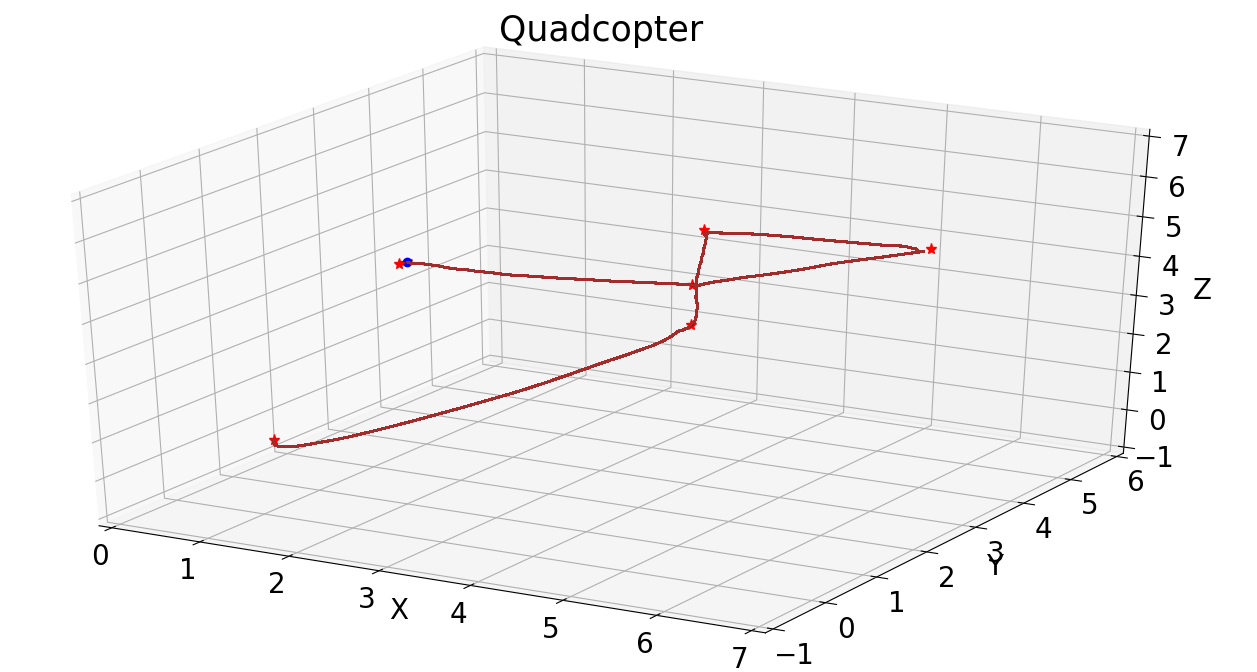}
	\caption{Quadcopter flight trajectory  based on full state observation } \label{Quadcopter_full_on}
\end{figure}
Figure (\ref{Quadcopter_full_on}) shows the flight trajectory based on full state observation. Compared to the trajectory in figure (\ref{Quadcopter_full_off}), the trajectory in figure (\ref{Quadcopter_full_on}) fluctuated in the initial stage.
\begin{figure}[t]
	\centering
	\includegraphics[width=\columnwidth]{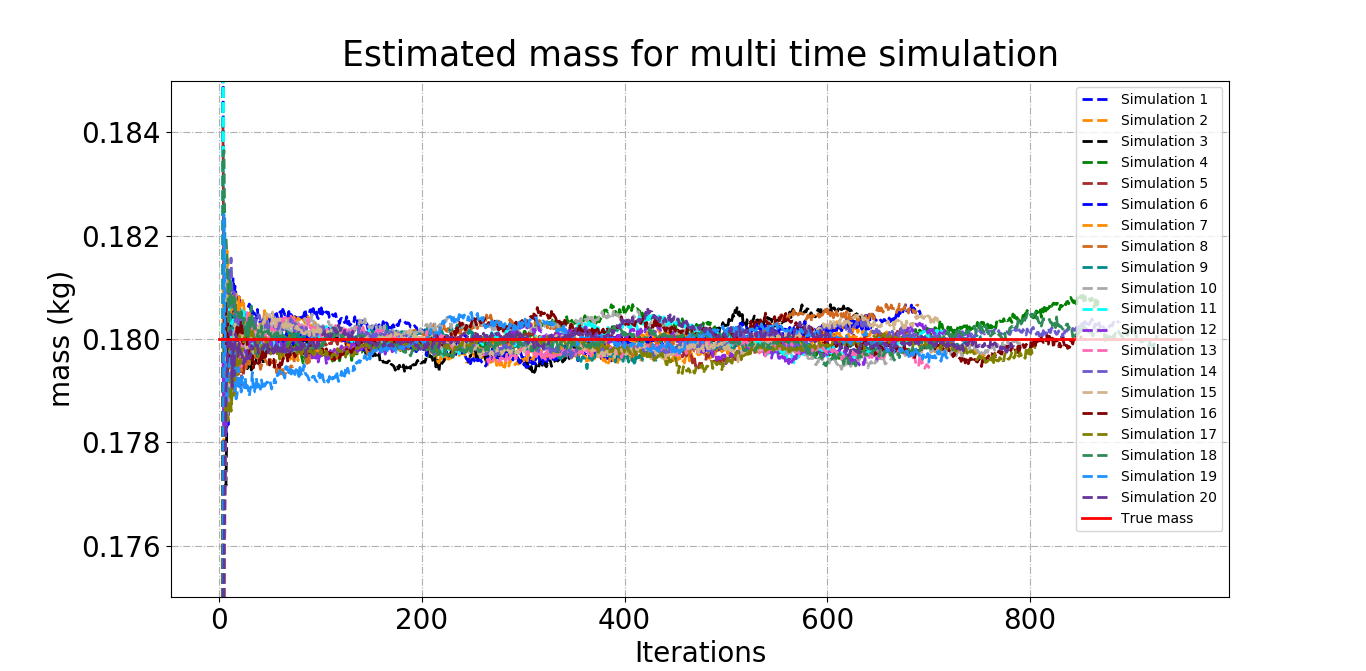}
	\caption{Estimated mass based on full state observation}\label{Estimated mass for multi time simulation_full_on}
\end{figure}
Figure (\ref{Estimated mass for multi time simulation_full_on}) shows the results of estimated mass based on full state observations. The red solid line is the true value of mass which is 0.18 kg. It can be seen that all the dashed lines gradually converged to the red solid line around 200th iteration. All the estimated mass from 20 simulation cycles is represented by dashed lines. The range of convergence value is (0.1793, 0.1807) kg. 
\begin{figure}[t]
	\centering
	\includegraphics[width=\columnwidth]{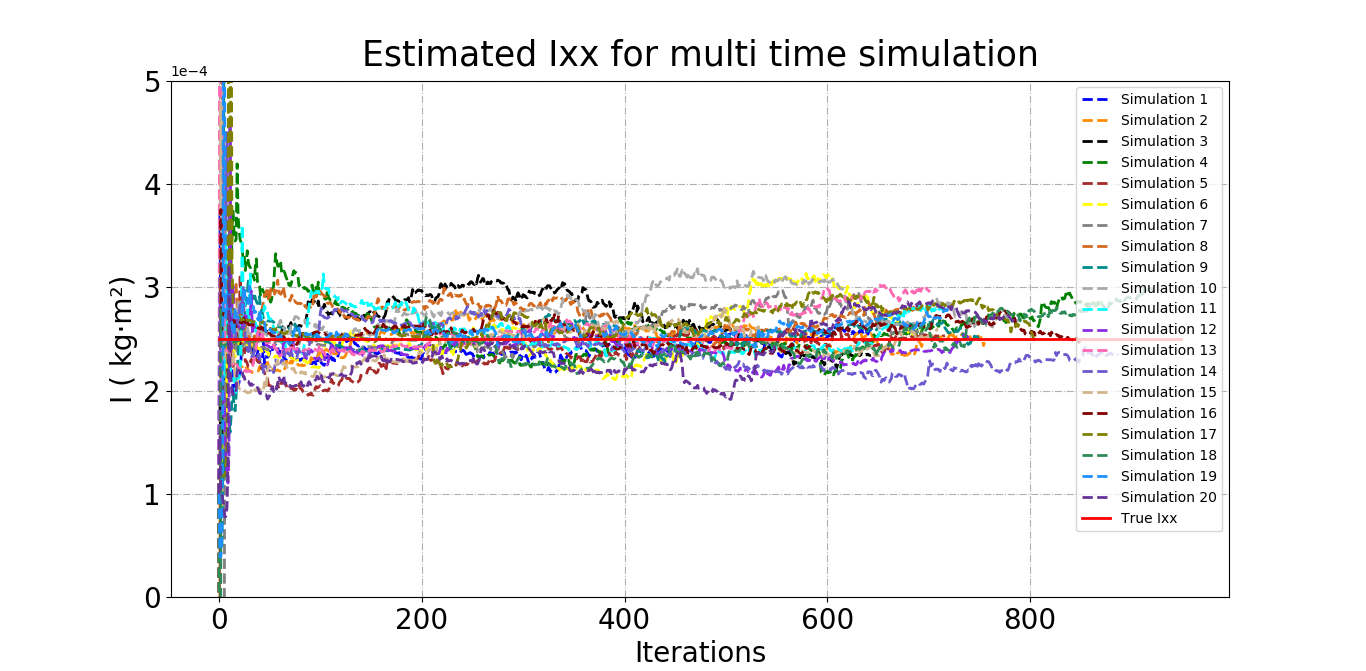}
	\caption{Estimated Ixx based on full state observation} \label{Estimated Ixx for multi time simulation_full_on}
\end{figure} 
The estimated $I_{xx}$ is shown in figure (\ref{Estimated Ixx for multi time simulation_full_on}). The true value of $I_{xx}$ is 2.5E-4 $\text{kg} \cdot \text{m}^{2}$ which is labeled by the red solid line. All the dashed lines distribute around the red solid line. All the dashed lines represent the results of estimated $I_{xx}$ based on 20 simulation cycles. The range of convergence value is (2E-4, 3.2E-4) $\text{kg} \cdot \text{m}^{2}$.  
\begin{figure}[t]
	\centering
	\includegraphics[width=\columnwidth]{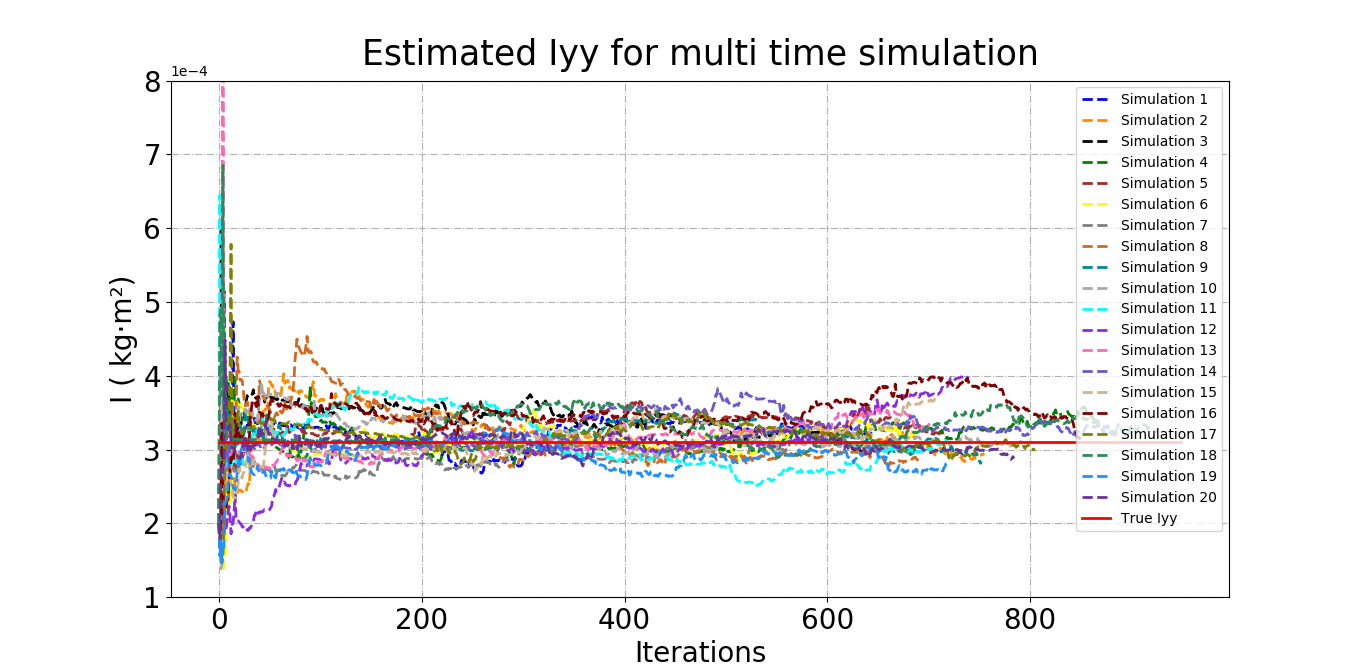}
	\caption{Estimated Iyy based on full state observation} \label{Estimated Iyy for multi time simulation_full_on}
\end{figure} 
\begin{figure}[t]
	\centering
	\includegraphics[width=\columnwidth]{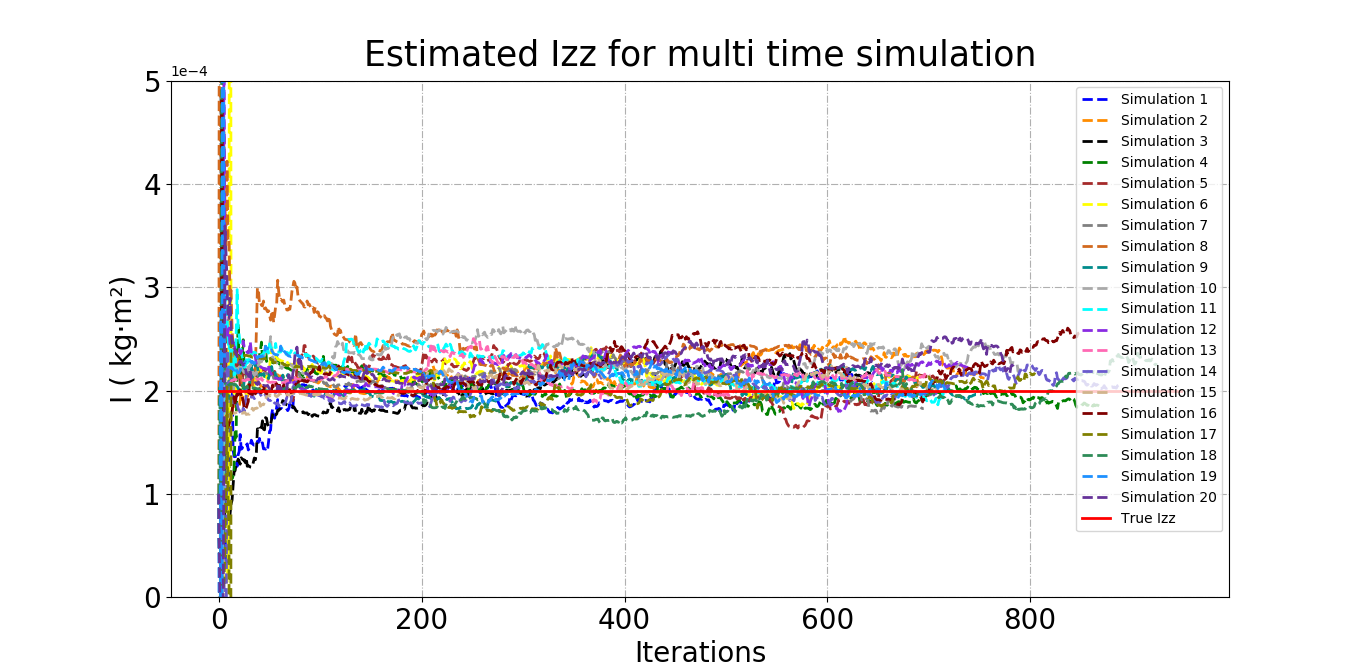}
	\caption{Estimated Izz based on full state observation} \label{Estimated Izz for multi time simulation_full_on}
\end{figure} 
Figure (\ref{Estimated Iyy for multi time simulation_full_on}) shows the results of estimated $I_{yy}$. It can be seen that all the dashed lines are gradually and stably distributed near the red solid line after 200 iterations. The solid red line is the true $I_{yy}$ 3.1E-4 $\text{kg} \cdot \text{m}^{2}$ and all dashed lines are the estimated results of $I_{yy}$. The range of convergence value is (2.5E-4, 4E-4) $\text{kg} \cdot \text{m}^{2}$. \\
Estimated $I_{zz}$ based on full state observation is shown in figure (\ref{Estimated Izz for multi time simulation_full_on}). All dashed lines fluctuate around the red solid line. The red solid line is true 2E-4 $\text{kg} \cdot \text{m}^{2}$. All dashed lines are the estimated $I_{zz}$ from 20 simulation cycles. The range of convergence value is (1.65E-4, 2.6E-4) $\text{kg} \cdot \text{m}^{2}$. 
\subsubsection{Based on partial state observation}
\begin{figure}[t]
	\centering
	\includegraphics[width=\columnwidth]{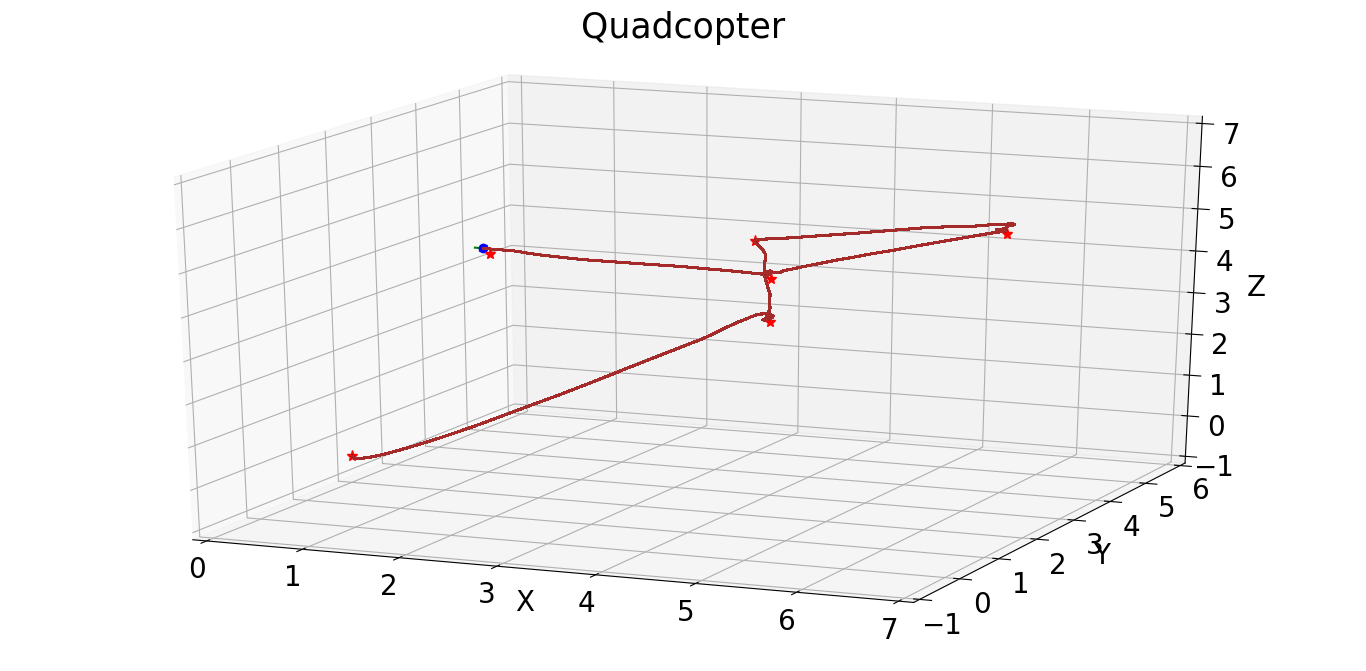}
	\caption{Quadcopter flight trajectory  based on partial state observation } \label{Quadcopter_part_on}
\end{figure}
Figure(\ref{Quadcopter_part_on}) is the flight trajectory based on partial state observation. This trajectory fluctuates slightly when reaching the first and third waypoints.
\begin{figure}[t]
	\centering
	\includegraphics[width=\columnwidth]{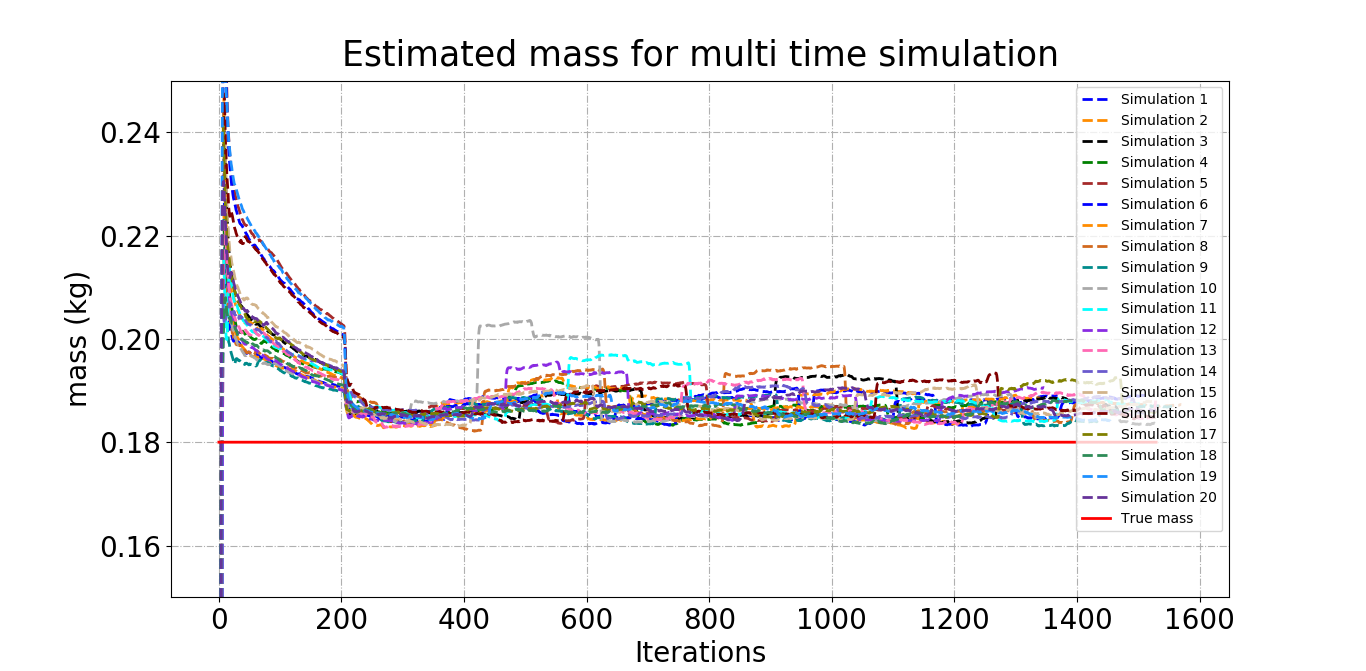}
	\caption{Estimated mass based on partial state observation}\label{Estimated mass for multi time simulation_part_on}
\end{figure}
Figure (\ref{Estimated mass for multi time simulation_part_on}) shows the estimated mass based on partial observation. It can be seen that all dashed lines gradually get close to the red solid line but are located above the red solid line. All dashed lines represent estimated mass. The red solid line represents true mass 0.18 kg. The range of convergence value of estimated mass is around (0.1825, 0.196) kg.
\begin{figure}[t]
	\centering
	\includegraphics[width=\columnwidth]{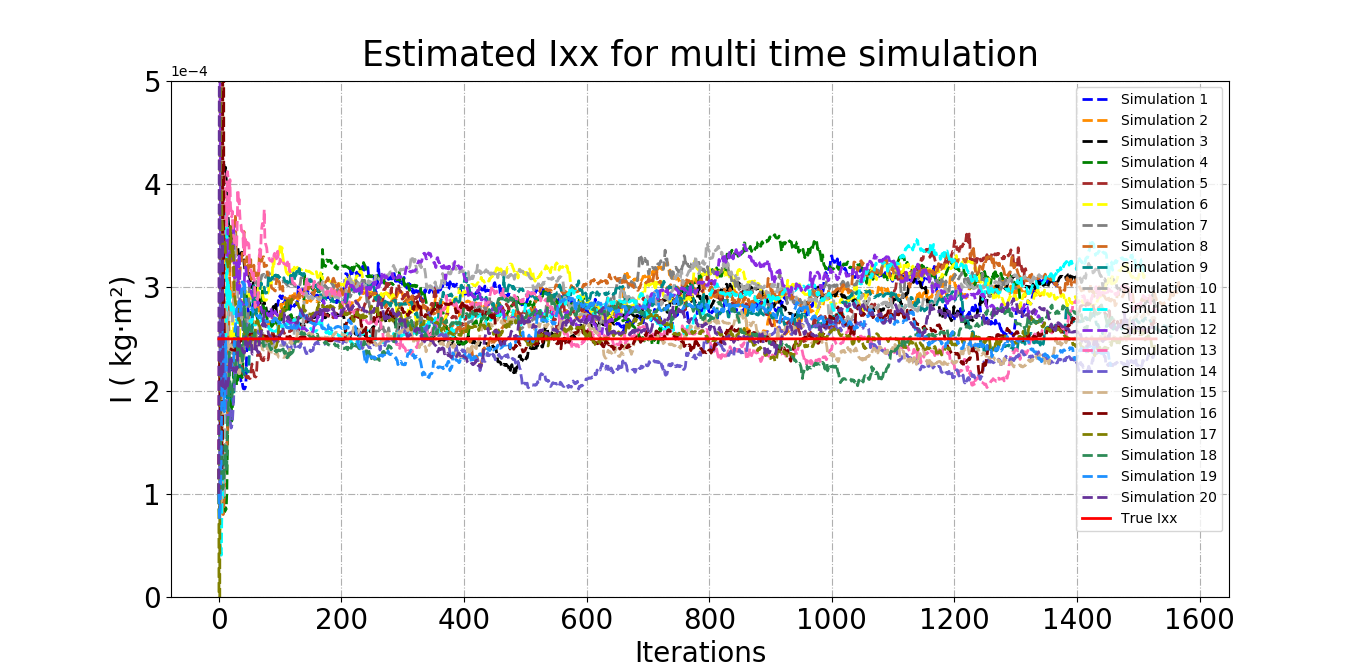}
	\caption{Estimated Ixx based on partial state observation} \label{Estimated Ixx for multi time simulation_part_on}
\end{figure} 
The estimated $I_{xx}$ results from 20 simulation cycles are shown in figure (\ref{Estimated Ixx for multi time simulation_part_on}). All dashed lines are distributed near the red solid line. The red solid line represents the true $I_{xx}$ 2.5E-4 $\text{kg} \cdot \text{m}^{2}$. All dashed lines represent the estimated $I_{xx}$. The range of convergence value of estimated $I_{xx}$ is around (2E-4, 3.5E-4) $\text{kg} \cdot \text{m}^{2}$.
\begin{figure}[t]
	\centering
	\includegraphics[width=\columnwidth]{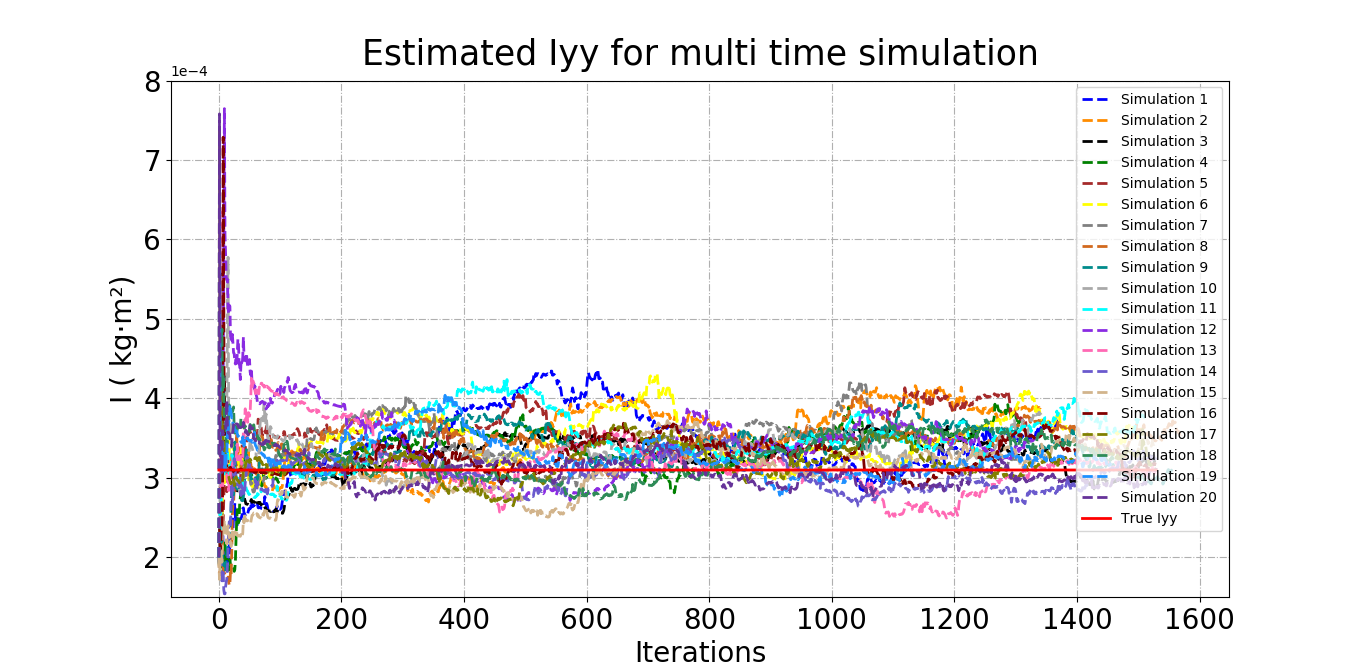}
	\caption{Estimated Iyy based on partial state observation} \label{Estimated Iyy for multi time simulation_part_on}
\end{figure} 
Figure (\ref{Estimated Iyy for multi time simulation_part_on}) shows the estimated $I_{yy}$ results from 20 simulation cycles. The dashed lines represent estimated $I_{yy}$ and the red solid line is true $I_{yy}$ 3.1E-4 $\text{kg} \cdot \text{m}^{2}$. It can be seen that dashed lines gradually converge around the red solid line. The range of the dashed line convergence value is around (2.5E-4, 4.2E-4) $\text{kg} \cdot \text{m}^{2}$.
\begin{figure}[t]
	\centering
	\includegraphics[width=\columnwidth]{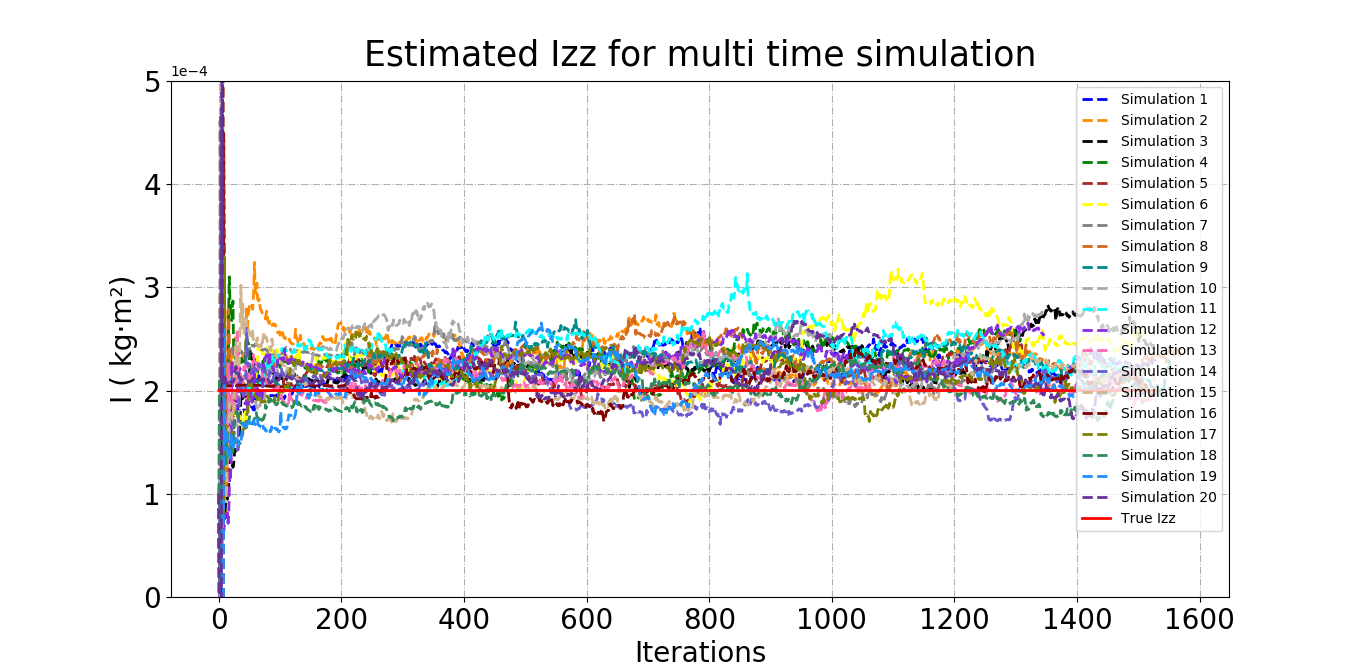}
	\caption{Estimated Izz based on partial state observation} \label{Estimated Izz for multi time simulation_part_on}
\end{figure} 
Estimated $I_{zz}$ results from 20 simulation cycles are shown in figure (\ref{Estimated Izz for multi time simulation_part_on}). The dashed lines in the figure represent estimated $I_{zz}$. The red solid line represents real $I_{zz}$ 2E-4 $\text{kg} \cdot \text{m}^{2}$. The dashed line in the figure undergoes short term fluctuations at the initial stage. After 200 iterations, it converges around the solid red line. The range of the dashed line convergence value is around (1.7E-4, 3.2E-4) $\text{kg} \cdot \text{m}^{2}$.
\subsubsection{Summary}
For better comparison, the data obtained above has been put into two tables. Table (\ref{Online estimated mass}) shows the estimated mass results based on three different types of observations sources. The second row of the table is the true mass. The third row to the fifth row represent the estimated mass convergence value range based on 20 simulation rounds. The estimated mass range based on EKF state estimation of online parameter estimation is approximately 2.2 times larger than that of offline parameter estimation. The estimated mass range based on full state observation of online parameter estimation is similar to that of offline parameter estimation. The estimated mass range based on partial state observation of online parameter estimation is approximately 2.1 times larger than that of offline parameter estimation. The parameter mass estimates based on partial state observations are distributed above the true values. The estimated mass is 1.39\% to 8.1\% higher than the real mass. One reason is that the Euler angle is not included in partial state observations.\\
\begin{table}[h]
	\renewcommand\arraystretch{1.5}
	\centering
	\setlength{\abovecaptionskip}{0pt}%
	\setlength{\belowcaptionskip}{10pt}%
	\caption{Online estimated mass }
	\begin{tabular}{c|c}
		\hline  
		& Mass (kg) \\ 
		\hline  
		Real value	 & 0.18  \\
		\hline 
		EKF state estimation  & (0.1795, 0.1806)   \\
		\hline 
		Full state observation	& (0.1793, 0.1807)   \\
		\hline 
		Partial state observation & (0.1825, 0.196)  \\
		\hline 
	\end{tabular}\label{Online estimated mass}
\end{table}
\begin{table*}[h]
	\renewcommand\arraystretch{1.5}
	\centering
	\setlength{\abovecaptionskip}{0pt}%
	\setlength{\belowcaptionskip}{10pt}%
	\caption{Online estimated inertia matrix}
	\begin{tabular}{p{3cm}|c|c|c}
		\hline  
		& $I_{xx} (\text{kg} \cdot  \text{m}^{2})$ & $I_{yy} (\text{kg} \cdot \text{m}^{2})$  & $I_{zz} (\text{kg} \cdot \text{m}^{2})$ \\ 
		\hline  
		Real value	  & 2.5E-4  & 3.1E-4	&2E-4\\
		\hline 
		EKF state estimation & (2E-4, 3.15E-4) & (2.65E-4, 3.75E-4) & (1.7E-4, 2.6E-4)  \\
		\hline 
		Full state observation	 & (2E-4, 3.2E-4)& (2.5E-4, 4E-4) & (1.65E-4, 2.6E-4)  \\
		\hline 
		Partial state observation & (2E-4, 3.5E-4)  & (2.5E-4, 4.2E-4) & (1.7E-4, 3.2E-4) \\
		\hline 
	\end{tabular}\label{Online estimated inertia matrix}
\end{table*}
Table (\ref{Online estimated inertia matrix}) shows the results of estimated inertia matrix based on three different types of observations sources. The second row of the table is the true value of inertial matrix components. The third row to the fifth row are the estimated inertial matrix components convergence range based on 20 simulation rounds. The second column of the table is $I_{xx}$, the third column is $I_{yy}$, and the fourth column is $I_{zz}$.\\
In $I_{xx}$ column, the convergence range of $I_{xx}$ estimation based on EKF state estimation is (2E-4, 3.15E-4) $\text{kg} \cdot  \text{m}^{2}$ and the convergence range of $I_{xx}$ estimation based on full state observation is (2E-4, 3.2E-4) $\text{kg} \cdot  \text{m}^{2}$. Compared with offline $I_{xx}$ estimates, the convergence range of $I_{xx}$ estimation based on EKF state estimation is around 3.46 times larger. This is due to the fact that in the offline parameter estimation module, the state observation of the entire trajectory is used for parameter estimation. The online parameter estimation module only uses the last 800 state observations for parameter estimation. However, the estimated convergence range of $I_{xx}$ based on partial state observations deviates to (2E-4, 3.5E-4) which is around 4.34 times larger than the range based on partial observations in offline parameter estimation. This is caused by the online parameter estimation module which has an impact on the controller and the number of observations which is less than the number of observations used in offline parameter estimation. The inertia matrix component $I_{xx}$ is related to torque $M$.\\
Similar to the estimation results of $I_{xx}$, the results of estimated $I_{yy}$ based on partial state observations also have a larger convergence range which is around 5.48 times larger than the range based on partial observations in offline parameter estimation. The deviation and enlargement of the convergence range are related to incomplete state observation and the number of state observations used for parameter estimation. \\
In $I_{zz}$ column, Compared with the convergence range of estimated $I_{zz}$ based on EKF state estimation and full state observation, the convergence range of estimated $I_{zz}$ based on partial state observations is relatively large. The estimated $I_{zz}$ convergence range based on partial estimation of online parameter estimation is approximately 5 times larger than that of offline parameter estimation.\\
In summary, the EM algorithm can efficiently realize parameter estimation. In addition, the number of state observations used for parameter estimation has an effect on parameter estimation. At the same time, the completeness of the state observation also has an effect on the parameter estimation. The effect of random noise on the system and observations can also be seen from the parameter estimation results.

\section{Conclusions and future work}
In this work, the effects of random noise are taken into account and a quadcopter based on the SDE model has been implemented. The LQG controller is added to the simulation model. Multiple simulations have shown that the LQG controller can effectively control the quadcopter. The state estimation module and the parameter estimation module have also been implemented. An expectation maximization algorithm has been used for parameter estimation. Offline parameter estimation simulation model and online parameter estimation simulation model have been implemented.\\
From the simulation results of the previous Section, the results of online parameter estimation are not as accurate as those of offline parameter estimation. The estimated inertia matrix convergence range of online parameter estimation is approximately 2.62 to 6 times larger than that of offline parameter estimation. The estimated mass convergence range of online parameter estimation based on EKF state estimation and partial state observation is approximately 2.1 to 2.2 times larger than that of offline parameter estimation. The effect of different types of observations on parameter estimation have been compared. Due to the lack of Euler angle observations, the estimated mass is higher than the real mass by 1.39\% to 8.89\%. In this paper, the results of 20 simulation cycles have been taken into account. Due to the uncertainty caused by random noise, the parameter convergence range in this paper is for reference only. The convergence range may also be slightly larger or smaller than the convergence range in this paper.\\
The work of this paper is limited to simulation. In the future, a real quadcopter can be used to verify the results shown in this paper. The controller can be improved and the small angle assumption can be removed. Other algorithms can be tried to estimate parameters in the future. Camera and roadblock recognition can also be added to the quadcopter.\\

\vspace{12pt}
\newpage
\appendix

\begin{equation*}
\begin{split}
\left[
\begin{array}{ccc}
\dot{x}\\
\dot{y}\\
\dot{z}\\
\ddot{x}\\
\ddot{y}\\
\ddot{z}\\
\dot{\phi}\\
\dot{\theta}\\
\dot{\psi}\\
\dot{\omega}_{x}\\
\dot{\omega}_{y}\\
\dot{\omega}_{z}
\end{array}
\right]_{k} =
\underbrace{
	\left[
	\begin{array}{cccccccccccc}
	0& 0 & 0 & 1 & 0 & 0 & 0 & 0 & 0 & 0& 0& 0\\
	0& 0 & 0 & 0 & 1 & 0 & 0 & 0 & 0 & 0& 0& 0\\
	0& 0 & 0 & 0 & 0 & 1 & 0 & 0 & 0 & 0& 0& 0\\
	0& 0 & 0 & 0 & 0 & 0 & 0 & 0 & 0 & 0& 0& 0\\
	0& 0 & 0 & 0 & 0 & 0 & 0 & 0 & 0 & 0& 0& 0\\
	0& 0 & 0 & 0 & 0 & 0 & 0 & 0 & 0 & 0& 0& 0\\ 
	0& 0 & 0 & 0 & 0 & 0 & 0 & 0 & 0 & 1& 0& 0\\
	0& 0 & 0 & 0 & 0 & 0 & 0 & 0 & 0 & 0& 1& 0\\
	0& 0 & 0 & 0 & 0 & 0 & 0 & 0 & 0 & 0& 0& 1\\
	0& 0 & 0 & 0 & 0 & 0 & 0 & 0 & 0 & 0& 0& 0\\
	0& 0 & 0 & 0 & 0 & 0 & 0 & 0 & 0 & 0& 0& 0\\
	0& 0 & 0 & 0 & 0 & 0 & 0 & 0 & 0 & 0& 0& 0
	\end{array}
	\right]}_{\mathbf{A}}
\left[
\begin{array}{ccc}
x\\
y\\
z\\
\dot{x}\\
\dot{y}\\
\dot{z}\\
\phi\\
\theta\\
\psi\\
\omega_{x}\\
\omega_{y}\\
\omega_{z}
\end{array}
\right]_{k} +
\underbrace{ 
	\left[
	\begin{array}{cccccc}
	0& 0& 0 & 0 & 0 & 0\\
	0& 0& 0 & 0 & 0 & 0\\
	0& 0& 0 & 0 & 0 & 0\\
	1& 0& 0 & 0 & 0 & 0\\
	0& 1& 0 & 0 & 0 & 0\\
	0& 0& 1 & 0 & 0 & 0\\
	0& 0& 0 & 0 & 0 & 0\\
	0& 0& 0 & 0 & 0 & 0\\
	0& 0& 0 & 0 & 0 & 0\\
	0& 0& 0 & 1 & 0 & 0\\
	0& 0& 0 & 0 & 1 & 0\\
	0& 0& 0 & 0 & 0 & 1
	\end{array}
	\right]}_{\mathbf{B}}
\underbrace{
	\left[
	\begin{array}{ccc}
	ua_{x}\\
	ua_{y}\\
	ua_{z}\\
	\dot{\omega}_{x}\\
	\dot{\omega}_{y}\\
	\dot{\omega}_{z}
	\end{array}
	\right]_{k}}_{u}
\end{split}
\end{equation*}

\begin{equation*}
\begin{split}
I_{xx} =  \frac{\sum_{k=2}^{n}\big[M_{x,k-1}^{2} + 2(I_{yy} - I_{zz})(M_{x,k-1}\omega_{y,k-1}\omega_{z,k-1}) + (I_{yy} - I_{zz})^{2}(\omega_{y,k-1}\omega_{z,k-1})^{2}\big]\triangle t}{\sum_{k=2}^{n} \big[(\omega_{x,k} - \omega_{x,k-1})[M_{x,k-1} + (I_{yy} - I_{zz})\omega_{y,k-1}\omega_{z,k-1}]\big] }
\end{split}\label{IxxEM}
\end{equation*}
Similarly, $I_{yy}$ and $I_{zz}$ can be derived as follows:\\
\begin{equation*}
\begin{split}
I_{yy} =  \frac{\sum_{k=2}^{n}\big[M_{y,k-1}^{2} + 2(I_{zz} - I_{xx})(M_{y,k-1}\omega_{x,k-1}\omega_{z,k-1}) + (I_{zz} - I_{xx})^{2}(\omega_{x,k-1}\omega_{z,k-1})^{2}\big]\triangle t}{\sum_{k=2}^{n} \big[(\omega_{y,k} - \omega_{y,k-1})[M_{y,k-1} + (I_{zz} - I_{xx})\omega_{x,k-1}\omega_{z,k-1}]\big] }
\end{split}
\end{equation*}
\begin{equation*}
\begin{split}
I_{zz} =  \frac{\sum_{k=2}^{n}\big[M_{z,k-1}^{2} + 2(I_{xx} - I_{yy})(M_{z,k-1}\omega_{x,k-1}\omega_{y,k-1}) + (I_{xx} - I_{yy})^{2}(\omega_{x,k-1}\omega_{y,k-1})^{2}\big]\triangle t}{\sum_{k=2}^{n} \big[(\omega_{z,k} - \omega_{z,k-1})[M_{z,k-1} + (I_{xx} - I_{yy})\omega_{x,k-1}\omega_{y,k-1}]\big] }
\end{split}
\end{equation*}
\end{document}